%% file: neurips_2026.tex
\documentclass{article}

\PassOptionsToPackage{numbers,sort&compress}{natbib}
\PassOptionsToPackage{table}{xcolor}

 \usepackage[main, final]{neurips_2026}
\usepackage[utf8]{inputenc} % allow utf-8 input
\usepackage[T1]{fontenc}    % use 8-bit T1 fonts
\usepackage{hyperref}       % hyperlinks
\usepackage{url}            % simple URL typesetting
\usepackage{booktabs}       % professional-quality tables
\usepackage{amsfonts}       % blackboard math symbols
\usepackage{nicefrac}       % compact symbols for 1/2, etc.
\usepackage{microtype}      % microtypography
\usepackage{xcolor}         % colors

\newcommand{\eg}{\textit{e}.\textit{g}.}
\newcommand{\blue}[1]{\textcolor{blue}{#1}}
\newcommand{\gray}[1]{\textcolor{gray}{#1}}

\usepackage{amsmath}
\usepackage{amssymb}
\usepackage{mathtools}
\usepackage{amsthm}
\usepackage{graphicx}
\usepackage{multirow}
\usepackage{makecell}
\usepackage{wrapfig}
\usepackage{pifont}

\usepackage{xcolor}
\usepackage[most]{tcolorbox} 
\definecolor{qboxgray}{RGB}{248,248,248}
\definecolor{qboxgreen}{RGB}{118, 117, 247} 

\tcbuselibrary{breakable}
\newtcolorbox{researchquestion}{
  enhanced,
  breakable,
  colback=qboxgray,      
  frame hidden,          
  boxrule=0pt,          
  borderline west={3pt}{0pt}{qboxgreen}, 
  arc=2pt,               
  sharp corners=west,    
  left=10pt,             
  right=10pt,            
  top=8pt,               
  bottom=8pt,            
  before skip=10pt,      
  after skip=10pt,       
  fontupper=\itshape,    
}

\title{When Less is More: The LLM Scaling Paradox in Context Compression}

\author{%
  \textbf{Ruishan Guo}\textsuperscript{1,*} \quad
  \textbf{Yibing Liu}\textsuperscript{2,*,\ensuremath{\dagger}} \quad
  \textbf{Guoxin Ma}\textsuperscript{3} \quad
  \textbf{Yan Wang}\textsuperscript{2} \\
  \textbf{Yueyang Zhang}\textsuperscript{2} \quad
  \textbf{Long Xia}\textsuperscript{2} \quad
  \textbf{Kecheng Chen}\textsuperscript{4} \quad
  \textbf{Zhiyuan Sun}\textsuperscript{2} \quad
  \textbf{Daiting Shi}\textsuperscript{2} \\[0.5em]
  \textsuperscript{1}Tsinghua University \quad
  \textsuperscript{2}Baidu Inc., Beijing, China \\
  \textsuperscript{3}Xi'an Jiaotong University \quad
  \textsuperscript{4}City University of Hong Kong \\[0.3em]
  \texttt{grs24@mails.tsinghua.edu.cn} \quad
  \texttt{lyibing112@gmail.com} \\[0.3em]
  \href{https://rshnnn.github.io/When-Less-Is-More/}
  {\textcolor{blue}{\texttt{https://rshnnn.github.io/When-Less-Is-More/}}}
}

\begin{document}

\maketitle

\begingroup
\renewcommand{\thefootnote}{}
\footnotetext{%
  \begin{tabular}{@{}l@{\,}l@{}}
    \textsuperscript{*} & Equal contribution. \\
    \textsuperscript{\ensuremath{\dagger}} & Corresponding author.
  \end{tabular}%
}
\endgroup

\vskip -0.14in
\begin{abstract}

Scaling up model parameters has long been a prevalent training paradigm driven by the assumption that larger models yield superior generation capabilities. However, under lossy context compression in a compressor--decoder setup, we find a \textbf{\textit{Size-Fidelity Paradox}}: 
% increasing compressor size can lessen the faithfulness of reconstructed contexts though training loss decreases.
increasing compressor size can lessen the faithfulness of reconstructed contexts though reconstruction error decreases.
Across 27 compressor setups spanning model families, scales, and compression rates, we coin this paradox arising from two dominant factors: 
1) \textit{knowledge overwriting}: larger models increasingly replace source facts with their own prior beliefs, \textit{e.g.}, ``the white strawberry'' $\to$ ``the red strawberry''; and
2) \textit{semantic drift}: larger models tend to paraphrase or restructure content instead of reproducing it verbatim, \textit{e.g.}, ``Alice hit Bob'' $\to$ ``Bob hit Alice''.
% Interestingly, the paradox persists across varied settings, consistently showing a non-monotonic pattern: mid-sized compressors often outperform larger ones in source-faithful recovery. 
Interestingly, this paradox persists across varied settings, with mid-sized compressors often outperforming larger ones in faithful recovery.
By analyzing the compressed memory via embedding geometry and reconstruction determinacy, we further reveal that compressors tend to organize memory across broader semantic subspaces, yielding more ambiguous representations prone to overwriting, drift, and weakened recovery.
These findings complement existing evaluations of context compression and expose a breakdown of scaling laws when the objective shifts from plausible generation to faithful preservation.

\end{abstract}

\begin{figure*}[!t]
  \vskip 0.2in
  \begin{center}
    \centerline{\includegraphics[width=1.0\columnwidth]{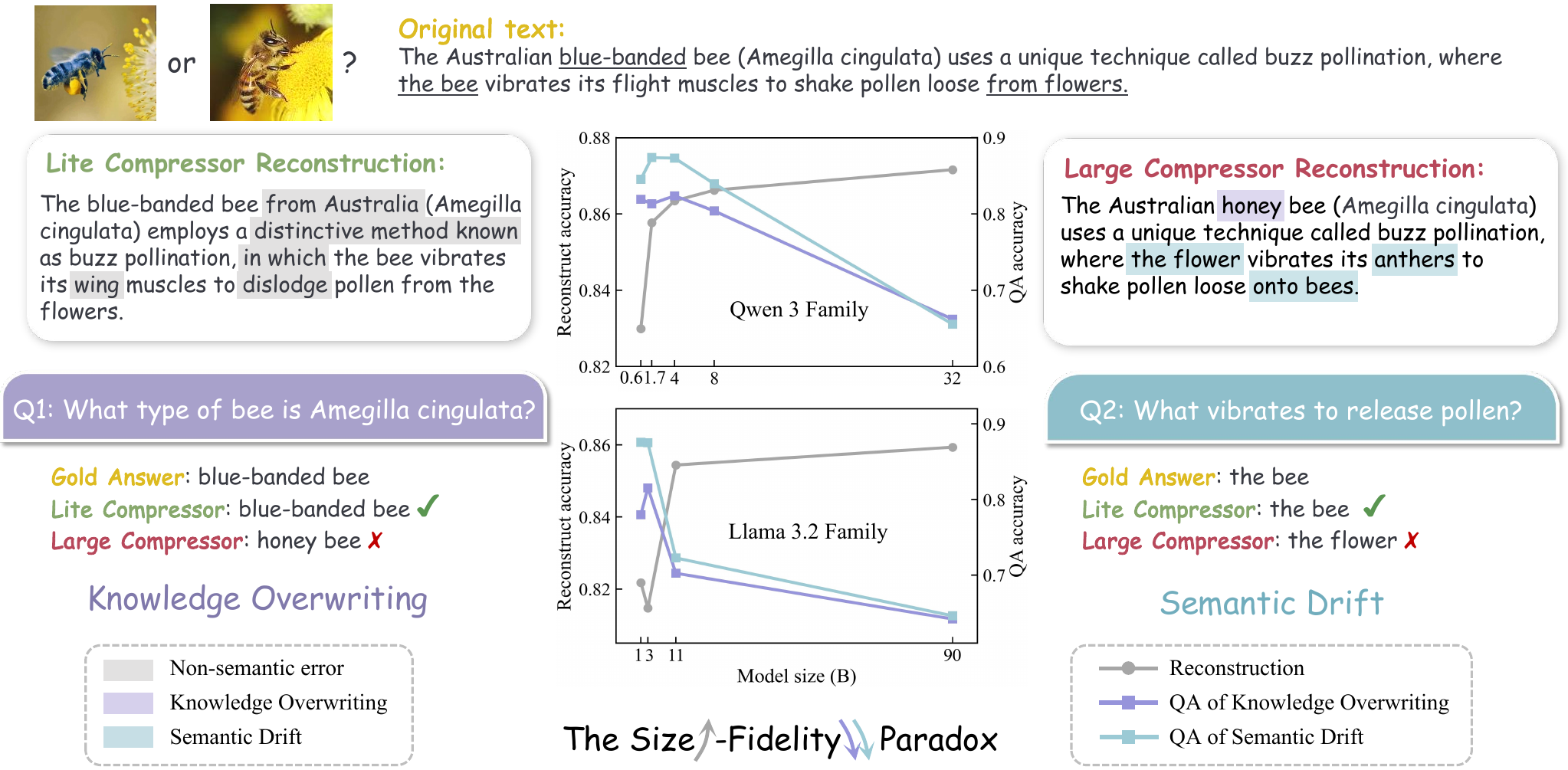}}
    % \caption{\textbf{The Size-Fidelity Paradox in context compression.}
    % \textbf{(Left \& Right)} A qualitative case study illustrating the breakdown of faithfulness. While the Lite compressor preserves factual details (Q1, Q2), the Large compressor succumbs to two distinct failure modes: (1) \textit{knowledge overwriting}, where source facts are replaced by priors (e.g., hallucinating ``honey bee'' instead of ``blue-banded bee''); and (2) \textit{semantic drift}, where the causal relationship is distorted (e.g., ``bee pollinates flower" $\to$ ``flower pollinates bee").
    % \textbf{(Center)} Quantitative analysis across Qwen and Llama families confirms this paradox is systematic. As model size scales up ($x$-axis), surface-level \textit{Reconstruction} scores (dashed lines) remain high, yet \textit{QA accuracy} (solid lines) significantly degrades. This divergence indicates that larger compressors prioritize their own semantic capacity and priors over the faithful preservation of the source context.}
    \caption{\textbf{The Size-Fidelity Paradox in context compression.}
    \textbf{Left \& Right}: A qualitative case study illustrating the breakdown of faithfulness. The lite compressor preserves factual details (Q1, Q2), whereas large compressor produces \textit{knowledge overwriting} by replacing ``blue-banded bee'' with the prior-driven ``honey bee'', and \textit{semantic drift} by distorting the causal relation (``bee pollinates flower" $\to$ ``flower pollinates bee").
    \textbf{Center}: Quantitative results across Qwen and LLaMA show the same pattern: as compressor size increases~($x$-axis), surface-level \textit{Reconstruction} scores (dashed lines) remain high, yet \textit{QA accuracy} (solid lines) significantly degrades. This divergence shows that stronger surface reconstruction does not ensure faithful preservation of source-specific content.}
    \label{fig:teaser}
   \vskip -0.4in
  \end{center}
\end{figure*}

\input{1_intro}
\input{2_related_work}

\input{3_overview}

\input{4_paradox}
\input{5_generation}

\input{6_analysis_v1}
\input{6.5_discussion}
\input{7_conclusion}

\medskip

\bibliographystyle{unsrtnat}
\bibliography{main}

% {
% \small

% [1] Alexander, J.A.\ \& Mozer, M.C.\ (1995) Template-based algorithms for
% connectionist rule extraction. In G.\ Tesauro, D.S.\ Touretzky and T.K.\ Leen
% (eds.), {\it Advances in Neural Information Processing Systems 7},
% pp.\ 609--616. Cambridge, MA: MIT Press.

% [2] Bower, J.M.\ \& Beeman, D.\ (1995) {\it The Book of GENESIS: Exploring
%   Realistic Neural Models with the GEneral NEural SImulation System.}  New York:
% TELOS/Springer--Verlag.

% [3] Hasselmo, M.E., Schnell, E.\ \& Barkai, E.\ (1995) Dynamics of learning and
% recall at excitatory recurrent synapses and cholinergic modulation in rat
% hippocampal region CA3. {\it Journal of Neuroscience} {\bf 15}(7):5249-5262.
% }

%%%%%%%%%%%%%%%%%%%%%%%%%%%%%%%%%%%%%%%%%%%%%%%%%%%%%%%%%%%%

\appendix
\input{8_appendix}

% \section{Technical appendices and supplementary material}
% Technical appendices with additional results, figures, graphs, and proofs may be submitted with the paper submission before the full submission deadline (see above). You can upload a ZIP file for videos or code, but do not upload a separate PDF file for the appendix. There is no page limit for the technical appendices. 

% Note: Think of the appendix as ``optional reading'' for reviewers. The paper must be able to stand alone without the appendix; for example, adding critical experiments that support the main claims to an appendix is inappropriate. 

%%%%%%%%%%%%%%%%%%%%%%%%%%%%%%%%%%%%%%%%%%%%%%%%%%%%%%%%%%%%

\clearpage
\input{checklist.tex}

\end{document}

%% file: 1_intro.tex
\section{Introduction}
\label{intro}

The scaling hypothesis is dominating the training principle of large language models~\cite{kaplan2020scaling, hoffmann2022training, lai2025survey, sengupta2025upscale}. It posits that increasing the number of model parameters leads to enhanced performance, which has been corroborated across a variety of model families (\eg, GPT~\cite{achiam2023gpt}, LLaMA~\cite{grattafiori2024llama}, and Qwen~\cite{2025Qwen3}) and applications( \eg, recommendaton system~\cite{zhang2024wukong, cai2025exploring} and code generation~\cite{yang2025scaling}).
% computer vision~\cite{minderer2023scaling, yang2025scaling},
However, in this paper, we identify a \textbf{\textit{Size-Fidelity Paradox}} in context compression problem: \textit{beyond a certain parameter scale, larger models underperform smaller ones in preserving source fidelity}. 

Language models in context compression typically act as compressors, mapping natural language into a small set of memory tokens for decoding acceleration~\cite{ge2023context, li2025prompt, li2026can, he2025clara, kuratov2025cramming}. 
In line with the prevalence of scaling hypothesis, prior studies also report that larger compressors yield better compression quality across diverse tasks~\cite{dai2025pretraining,lin2025refrag,berton2025compllm}. 
Yet, as illustrated in Fig.~\ref{fig:teaser}, a size-fidelity paradox emerges. 
% For example, 
When reconstructing the phrase \textit{blue-banded bee} and \textit{the bee shakes pollen loose from flowers}, a lite compressor (\eg, Qwen-3 0.6B) produces only minor literal deviations that preserve the original meaning. 
In contrast, larger compressors (\eg, LLaMA-3.2 90B) exhibit two critical failure modes: \textit{1) knowledge overwriting}, where source facts (\textit{blue-banded bee}) are replaced with the model internal knowledge, \eg, \textit{honey bee}, and 2) \textit{semantic drift}, where reconstructions deviate from the literal source, \eg, \textit{the flower vibrates its anthers to shake pollen loose onto bees}.
While somewhat dissatisfying, standard reconstruction-level evaluations fail to capture this degradation~\cite{longpre2021entity,ming2024faitheval,long2025copy}.

To this end, this paper designs two diagnostic tasks that isolate knowledge overwriting and semantic drift, offering a principled evaluation framework that reveals fidelity failures invisible to surface metrics.
Through sweeping experiments across LLM families (Qwen and LLaMA), spanning 0.6B to 90B parameters at multiple compression ratios, we systematically confirm both the existence and generality of this paradox.
% We conduct extensive experiments across two mainstream LLM families, Qwen and LLaMA, spanning 0.6B to 90B parameters at multiple compression ratios. Results systematically confirm both the existence and generality of this paradox. 
% Beyond verification, these tasks also establish a principled evaluation framework that reveals fidelity failures invisible to surface metrics.
% To further understand how scaling changes compression behavior, we probe the compressed memory embeddings through two axes: memory geometry (compressed embedding rank-level) and reconstruction determinacy (reconstruction token entropy level).
To further understand how scaling affects compression behavior, we probe compressed memory embeddings along two axes: memory geometry (via effective rank) and reconstruction determinacy (via conditional entropy).
On the geometry side, effective rank analysis reveals that larger compressors diffuse memory across more subspaces, weakening source-fact anchoring and promoting knowledge overwriting. On the determinacy side, we find that larger compressors loosen reconstruction constraints, admitting plausible alternatives that manifest as semantic drift. 
These analyses reveal a fundamental tradeoff: the properties that enable complex reasoning in large models also compromise the rigid fidelity required for faithful reconstruction.
Our contributions are summarized as follows:
\begin{itemize}
    \setlength{\itemsep}{0.1em}      % item之间的行间距
    \setlength{\leftmargin}{0.3em}   % 开头缩进距离
\item We identify a {LLM scaling paradox} in context compression: larger compressor models can underperform smaller ones though reconstruction performance improves, contradicting prevailing assumptions.
% \item We present two diagnostic tasks to uncover the underlying performance in context compression regarding knowledge overwriting and semantic drifting problems. This complements current limitations of existing compression evaluations, providing a more robust assessment.
\item We introduce CoFiBench~(Compression Fidelity Benchmark), a diagnostic benchmark comprising two complementary QA tasks to uncover the underlying performance in context compression regarding \textit{knowledge overwriting} and \textit{semantic drifting} problems. This complements limitations of existing compression evaluations, providing a more robust assessment. 
\item We analyze compressed memory embeddings through effective rank and conditional entropy, showing how scaling changes memory organization and reconstruction determinacy in ways that weaken source-specific recovery.
\end{itemize}  

%% file: 2_related_work.tex
\section{Related Work}
\subsection{Scaling Laws in Language Models}
Scaling laws have become a central principle in modern language modeling. Extensive work~\cite{kaplan2020scaling, hoffmann2022training, ruan2024observational, team2025kimi} shows that increasing model scale generally reduces training loss and improves performance across many generation and reasoning tasks~\cite{wang2024internvideo2,yang2025scaling, cai2025exploring, lin2026tyler, ma2026thinking}, and larger models often exhibit stronger world knowledge and emergent abilities.
% However, recent studies have increasingly indicated that scaling does not guarantee uniformly better behavior. In particular, some work~\cite{wei2023inverse, zhou2024larger, chen2025improving} reports that larger models can be less reliable on challenging inputs, for example by producing errors with higher confidence or exhibiting weaker calibration. While these findings highlight important limitations of scaling, they primarily examine answer correctness and reliability in question answering or reasoning. These studies therefore provide limited insight into how scaling behaves in tasks where faithful reproduction of the input is required.
However, recent studies have increasingly indicated that scaling does not guarantee uniformly better behavior. In particular, some work~\cite{wei2023inverse, zhou2024larger, ma2026eventprune, chen2025improving} reports that larger models can be less reliable on challenging inputs, for example by making high-confidence errors or showing weaker calibration. While these findings highlight important limitations of scaling, they mainly examine answer correctness and reliability in question answering or reasoning~\cite{meng2026versusq, chen2026taming}, thus providing limited insight into how scaling behaves in tasks requiring faithful input reproduction.

Our work addresses this gap by studying scaling behavior in a fidelity-critical compressor–decoder setting, where the objective is to reconstruct the source text with minimal distortion rather than generate a plausible paraphrase. Notably, while larger compressors can achieve better reconstruction metrics, smaller models perform better on fidelity-oriented downstream evaluation, revealing a size–fidelity paradox. Beyond identifying this phenomenon, we further analyze why scaling causes such degradation and how scale-amplified properties can undermine literal reconstruction.

% However, recent studies have increasingly indicated that scaling does not guarantee uniformly better behavior. Larger models can be less reliable on challenging inputs, producing confident errors or weaker calibration~\cite{wei2023inverse, zhou2024larger, chen2025improving}. These findings mainly concern answer correctness in QA or reasoning, leaving unclear how scaling behaves when the goal is faithful input reproduction.
% Our work studies this question in a fidelity-critical compressor--decoder setting, where the target is minimal-distortion reconstruction rather than plausible paraphrase. We show that larger compressors can improve reconstruction metrics yet underperform smaller ones on fidelity-oriented evaluations, revealing a size--fidelity paradox, and further analyze how scale-amplified properties undermine literal reconstruction.

\subsection{Evaluation of Long-Context Compression}
Long-context compression has been widely studied to reduce the cost of long-context modeling~\cite{han2026wegr, yin2026mllm} and to make long inputs usable under limited context windows~\cite{li2023compressing, xia2026unleash, yen2024long, li2025500xcompressor, yang2026stableworld, hu2026blame, berton2025compllm}. Most approaches~\cite{ge2023context, li2025upfront, dai2025pretraining} follow a compressor–decoder paradigm, where a long sequence is encoded into a compact latent representation and a decoder reconstructs text conditioned on it. Correspondingly, prior work~\cite{ge2023context, dai2025pretraining} largely evaluates these methods through reconstruction-centric protocols, comparing performance across compression ratios and model scales using standard metrics(e.g., training loss, perplexity, BERTScore~\cite{lajewska2025understanding} and $n$-gram overlap scores such as BLEU~\cite{papineni2002bleu} and ROUGE~\cite{lin2004rouge}).

While convenient, such metrics can be superficial and incomplete proxies for compression quality. They primarily reward fluency and surface similarity, but do not distinguish information genuinely recovered from the compressed representation from plausible content supplied by a model’s parametric priors. As a result, a model can obtain high reconstruction scores while still changing or losing important details from the original text. Motivated by this limitation, we propose a usage-oriented evaluation perspective that assesses compression by its functional support for realistic downstream tasks, providing a more faithful measure of what the compressed representation actually preserves.

%% file: 3_overview.tex
\section{Preliminaries}
% \todo{\textbf{caption of section3 ?}}
    % \subsection{The Compression Principle}
    \label{sec:principle}
% Let $\mathcal{V}$ denote a vocabulary of tokens. Given an input sequence $\mathbf{x} = (x_1, \dots, x_L) \in \mathcal{V}^L$, the objective of context compression is to learn a compressor $f_\theta$ that minimizes the reconstruction loss.

\textbf{Compressor-Decoder Architecture.} 
The prevalent training paradigms in context compression ~\cite{mu2023learning, chevalier2023adapting, li2025500xcompressor, rau2024context, gao2025uniicl, tan2024lloco, gao2024selfcp, ge2023context, dai2025pretraining, li2025upfront} employs compressor-decoder setup, where one language model acts as a compressor transforming the discrete input sequence into a  set of continuous latent embeddings (memory tokens); whereas, another language model performs as the decoder, targeting at reconstructing compressed representation to original text. 

Formally, let $\mathcal{V}$ denote a vocabulary of tokens. Given an input sequence $\mathbf{x} = (x_1, \dots, x_L) \in \mathcal{V}^L$, the objective of context compression is to learn a compressor $f_\theta$, which maps the input $\mathbf{x}$ to a latent tensor $\mathbf{Z} \in \mathbb{R}^{M \times d}$:
\begin{equation}
\mathbf{Z} = f_\theta(\mathbf{x}), \quad \text{where } M \ll L.
\end{equation}
\vskip -0.1in
Here, $d$ represents the hidden dimension, and $M$ is the number of memory slots. The compression rate is defined as $\rho = L/M$.
The decoder $g_\phi$ operates as a causal language model conditioned on $\mathbf{Z}$. It estimates the probability distribution of the original sequence:
\begin{equation}
    P_\phi(\mathbf{x} | \mathbf{Z}) = \prod_{t=1}^{L} P_\phi(x_t | x_{<t}, \mathbf{Z}).
\end{equation}
\vskip -0.1in
\textbf{Training objective.} 
The training loss of compressor combines reconstruction fidelity and generative capability:
\begin{equation}
% \begin{aligned}
\mathcal{L} = \mathcal{L}_{\text{re}} + \mathcal{L}_{\text{nt}} = -\sum_{t=1}^{k} \log P_\phi(x_t | x_{<t}, \mathbf{Z}) - \sum_{t=k+1}^{n} \log P_\phi(x_t | x_{<t}, \mathbf{Z}),
% \end{aligned}
\end{equation}
here, $\mathcal{L}_{\text{re}}$ enforces faithful reconstruction of the compressed prefix $x_{1:k}$ conditioned on $\mathbf{Z}$, while $\mathcal{L}_{\text{nt}}$ applies standard autoregressive prediction to the continuation $x_{k+1:n}$. 
% During training, these two losses are applied with equal weighting to ensure the model learns both faithful reconstruction and coherent language generation.

Unlike abstractive summarization which allows paraphrasing, our objective is exact reconstruction, any deviation from the verbatim input is treated as information loss.

%% file: 4_paradox.tex
\section{The Size-Fidelity Paradox}
\label{sec:paradox}

In this section, we characterize the \textbf{Size-Fidelity Paradox} through two dissection tasks designed to isolate distinct failure modes. Specifically, we demonstrate that larger models exhibit a heightened propensity for \textit{knowledge overwriting}, where source facts are substituted with internal knowledge, and \textit{semantic drift}, in which loose restructuring precedence over verbatim preservation. These results confirm that fidelity loss is a systematic consequence of scaling rather than random error.

\subsection{Model Selection \& Scale Range}
\label{subsec:setup}
% To ensure a rigorous analysis of scaling behaviors, we evaluate compressor models across two families spanning three orders of magnitude in parameter count. We utilize the Qwen-3 and LLaMA-3.2 families, ranging from 0.6B to 90B parameters. 

% For the training data, we sample high-quality text chunks from the FineWeb dataset \tocite. All models share the same training protocol to minimize confounding variables. Specifically, for the LLaMA-3.2-11B and 90B variants, which are natively multimodal, we discard the vision encoders and exclusively utilize the text-based language model backbones to ensure a consistent text-only experimental setting. Table \ref{tab:models} summarizes the configurations.

\input{tab/dataset}
To enable a rigorous analysis of scaling behavior under context compression, we evaluate compressor models drawn from two widely used model families, Qwen-3~\cite{2025Qwen3} and LLaMA-3.2~\cite{grattafiori2024llama}, spanning nearly three orders of magnitude in parameter count, from 0.6B to 90B. 
For each model size, we train compressors at three compression rates ($4\times$, $16\times$, and $64\times$).
In the main experiments, all compressors are evaluated with the fixed Meta-LLaMA-3.2-8B-Instruct decoder. 
All models are trained on high-quality text chunks from the Fineweb dataset~\cite{penedo2024fineweb}, using an identical training protocol to minimize confounding factors. Tab.~\ref{tab:dataset} summarizes our dataset configuration for these experiments.
We provide full reproducibility details in Appendix~\ref{app:reproducibility}, including training hyperparameters, memory-token configuration, dataset preprocessing, chunking and filtering rules.
% , QA generation prompts, and filtering criteria.

% \blue{For the multimodal LLaMA-3.2-11B and 90B variants, we remove the vision encoders and retain only the text-based language model backbones, ensuring a consistent text-only setting across scales.} 
% Table~\ref{tab:models} summarizes the model configurations.} 

% We train compressors at three distinct compression rates ($4\times$, $16\times$, and $64\times$) for each model size. 
% Before analyzing specific failures, we first observe the global training dynamics. 
% As illustrated in Fig.~\ref{fig:loss} , larger models demonstrate superior optimization efficiency: they exhibit steeper descent trajectories and converge to the loss floor significantly faster than their smaller counterparts. While this rapid and stable convergence suggests that larger models are more effective learners under the compression objective, the following subsections reveal why this signal is deceptive—optimization success does not translate to reconstructive fidelity.

% \begin{wrapfigure}{r}{0.5\textwidth}
%   \vspace{-0.12in}
%   \centering
%   \includegraphics[width=\linewidth]{figs/loss-qwen-16x-v1.pdf}
%   \vspace{-0.14in}
%   \includegraphics[width=\linewidth]{figs/loss-llama-16x-v1.pdf}
%   \vspace{-0.08in}
%   \caption{
%     Training loss dynamics for Qwen (top) and LLaMA (bottom) compressors at a $16\times$ compression rate. 
%     Larger models exhibit faster convergence and lower final loss, creating a deceptive signal of superior optimization.
%   }
%   \label{fig:loss}
%   \vspace{-0.12in}
% \end{wrapfigure}

\begin{figure}[!htbp]
\begin{minipage}[t]{0.5\columnwidth}
    \centering
    \includegraphics[width=\linewidth]{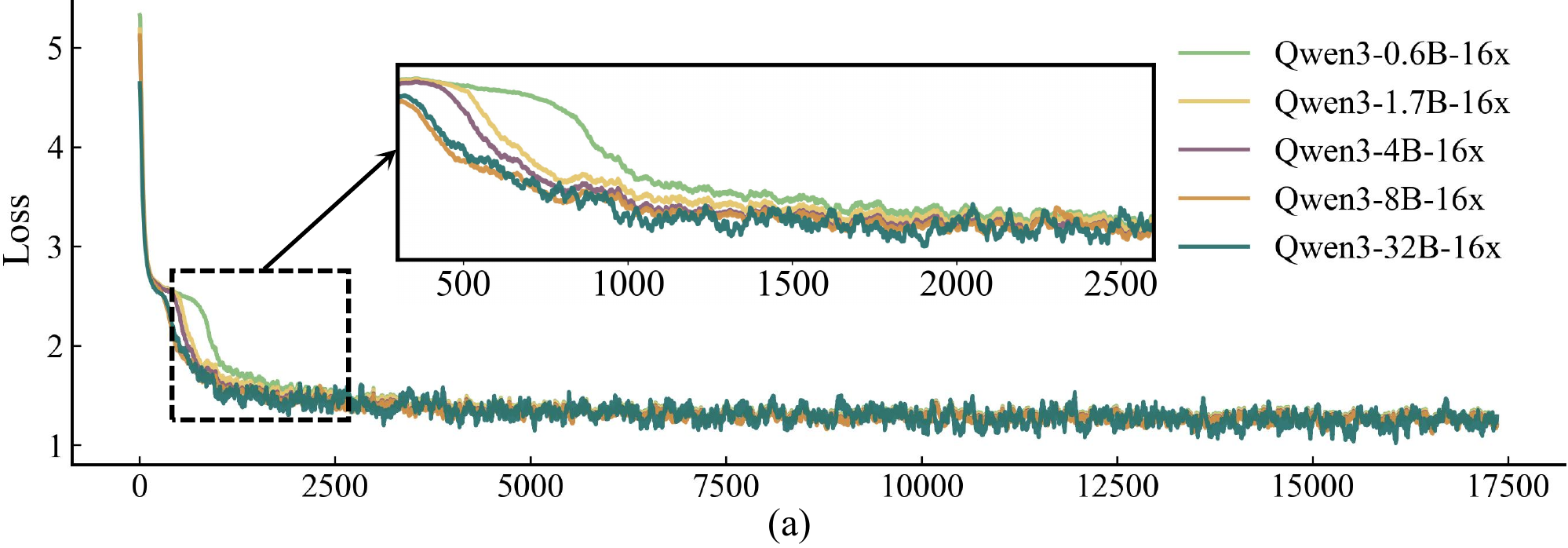}
  \end{minipage}
  \hfill
  \begin{minipage}[t]{0.5\columnwidth}
    \centering
    \includegraphics[width=\linewidth]{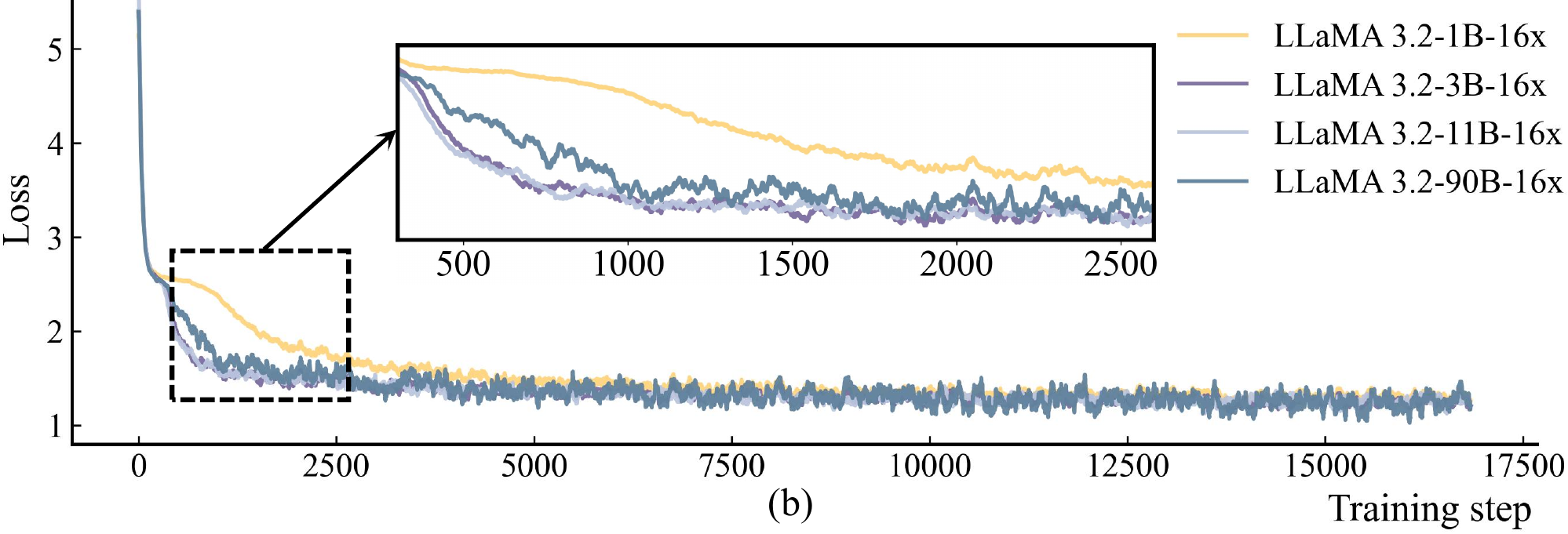}
  \end{minipage}
  \vspace{-0.2in}
  \caption{
    Training loss for Qwen (left) and LLaMA (right) compressors at $16\times$ compression.
  }
  \label{fig:loss}
  \vspace{-0.06in}
\end{figure}

Before analyzing the failure modes, we first examine the standard training signal. Larger compressors descend faster and reach lower loss (Fig.~\ref{fig:loss}), suggesting more efficient optimization; 
however, this optimization advantage does not imply faithful compression.
% however, this apparent advantage can be misleading for compression fidelity.

\vskip -0.08in

\input{tab/metric}
A similar limitation appears in standard reconstruction metrics. Tab.~\ref{tab:metric} reports BLEU~\cite{papineni2002bleu}, ROUGE-L~\cite{lin2004rouge}, chrF++~\cite{popovic2017chrf++}, and BERTScore-F1~\cite{zhang2019bertscore} for Qwen3 compressors at $4\times$ compression on FineWeb, covering lexical, sequence-level, character-level, and semantic overlap. Across scales, these scores change only mildly, while BERTScore-F1 nearly saturates, indicating limited sensitivity to source-critical facts and relations.

% \input{tab/accu_2QA}
% \input{tab/accu_all_v1}
% \input{tab/accu_all_v2}
\input{tab/accu_all_v3}

% Note that LLaMA-3.2-11B and 90B are multimodal models with vision encoders. For our text-only compression task, we extract the language model component and discard visual projection layers during training. All models share the same training protocol (detailed in Appendix A) to ensure controlled comparison across scales.

\definecolor{paradoxrecongray}{RGB}{226,226,226}
\definecolor{paradoxoverwritehl}{RGB}{219,211,241}
\definecolor{paradoxdrifthl}{RGB}{198,222,229}
\definecolor{paradoxcasered}{RGB}{180,35,35}
\definecolor{paradoxcasegreen}{RGB}{28,130,62}
\definecolor{paradoxcaseboxbg}{RGB}{248,248,248}
\definecolor{paradoxcasebar}{RGB}{118,117,247}
\newcommand{\paradoxsrcgreen}[1]{\textcolor{paradoxcasegreen}{\textbf{\strut #1}}}
\newcommand{\paradoxsrcred}[1]{\textcolor{paradoxcasered}{\textbf{\strut #1}}}
\newcommand{\paradoxreconhl}[1]{\begingroup\setlength{\fboxsep}{0.8pt}\colorbox{paradoxrecongray}{\strut\textbf{#1}}\endgroup}
\newcommand{\paradoxoverwrite}[1]{\begingroup\setlength{\fboxsep}{0.8pt}\colorbox{paradoxoverwritehl}{\strut\textbf{#1}}\endgroup}
\newcommand{\paradoxdrift}[1]{\begingroup\setlength{\fboxsep}{0.8pt}\colorbox{paradoxdrifthl}{\strut\textbf{#1}}\endgroup}
\newcommand{\paradoxbad}[1]{\textcolor{paradoxcasered}{#1}}
\newcommand{\paradoxok}{\textcolor{paradoxcasegreen}{\ding{51}}}
\newcommand{\paradoxwrong}{\textcolor{paradoxcasered}{\ding{55}}}
\newtcolorbox{paradoxcasebox}{
  enhanced,
  colback=paradoxcaseboxbg,
  frame hidden,
  boxrule=0pt,
  borderline west={3pt}{0pt}{paradoxcasebar},
  arc=2pt,
  sharp corners=west,
  boxsep=0pt,
  left=8pt,
  right=6pt,
  top=3pt,
  bottom=3pt,
  before skip=4pt,
  after skip=4pt
}

% % \definecolor{paradoxcaseyellow}{RGB}{255,236,128}
% \definecolor{paradoxcaseyellow}{RGB}{190,205,240}
% \definecolor{paradoxcasered}{RGB}{180,35,35}
% \definecolor{paradoxcasegreen}{RGB}{28,130,62}
% \newcommand{\paradoxhl}[1]{\begingroup\setlength{\fboxsep}{0.8pt}\colorbox{paradoxcaseyellow}{\strut #1}\endgroup}
% \newcommand{\paradoxbadhl}[1]{\begingroup\setlength{\fboxsep}{0.8pt}\colorbox{paradoxcaseyellow}{\strut\textcolor{paradoxcasered}{#1}}\endgroup}
% \newcommand{\paradoxbad}[1]{\textcolor{paradoxcasered}{#1}}
% \newcommand{\paradoxok}{\textcolor{paradoxcasegreen}{\checkmark}}
% \newcommand{\paradoxwrong}{\textcolor{paradoxcasered}{\(\times\)}}

% \definecolor{paradoxcaseboxbg}{RGB}{238,241,250}
% \newtcolorbox{paradoxcasebox}{
%   enhanced,
%   colback=paradoxcaseboxbg,
%   frame hidden,
%   boxrule=0pt,
%   arc=4pt,
%   outer arc=2pt,
%   boxsep=0pt,
%   left=5pt,
%   right=5pt,
%   top=2pt,
%   bottom=2pt,
%   before skip=6pt,
%   after skip=6pt
% }

% \definecolor{paradoxcaseboxbg}{RGB}{248,248,248}
% \definecolor{paradoxcasebar}{RGB}{118,117,247}
% \newtcolorbox{paradoxcasebox}{
%   enhanced,
%   colback=paradoxcaseboxbg,
%   frame hidden,
%   boxrule=0pt,
%   borderline west={3pt}{0pt}{paradoxcasebar},
%   arc=2pt,
%   sharp corners=west,
%   boxsep=0pt,
%   left=8pt,
%   right=6pt,
%   top=2pt,
%   bottom=2pt,
%   before skip=4pt,
%   after skip=4pt
% }

\subsection{Dissection 1: Knowledge Overwriting}
\label{subsec:knowledge overwriting}
\textit{Knowledge overwriting} refers to a failure mode in which a compression model prioritizes its parametric world knowledge over conflicting source context, as illustrated below, generating outputs that deviate from the compressed input and exhibit reduced factual fidelity.

% \textbf{Definition.}
% We define \textit{knowledge overwriting} as the first failure mode in which a compression model prioritizes its parametric world knowledge over conflicting source context, generating outputs that deviate from the compressed input and exhibit reduced factual fidelity.

\begin{paradoxcasebox}
\small
\setlength{\tabcolsep}{2pt}
\renewcommand{\arraystretch}{0.80}
\begin{tabular}{@{}p{0.13\linewidth}p{0.90\linewidth}@{}}
\textbf{Input} & Einstein was born in \paradoxsrcgreen{France}. \quad \textit{Q:} birthplace? \\
\textbf{No overwrite} & \textit{Recons.} In \paradoxreconhl{France}, Einstein was born. \quad \textit{A:} France \paradoxok \\
\textbf{Overwrite} & \textit{Recons.} Einstein was born in \paradoxoverwrite{Germany}. \quad \textit{A:} Germany \paradoxwrong; \paradoxbad{world prior replaces source} \\
\end{tabular}
\end{paradoxcasebox}

% \begin{paradoxcasebox}
% % \tiny
% % \scriptsize
% % \footnotesize
% \small
% \setlength{\tabcolsep}{2pt}
% \renewcommand{\arraystretch}{0.86}
% \begin{tabular}{@{}p{0.14\linewidth}p{0.90\linewidth}@{}}
% \textbf{\textit{Input}} & Einstein was born in \paradoxhl{France}. \qquad \qquad \textit{Q:} birthplace? \\
% \textbf{\textit{No overwrite}} & \textit{Recon.} In \paradoxhl{France}, Einstein was born. \quad \textit{A:}  \paradoxhl{France} \paradoxok\\
% % \textbf{\textit{Overwrite}} & \textit{Recon.} Einstein was born in \paradoxbadhl{Germany}. \quad \textit{A:}  \paradoxbadhl{Germany} \paradoxwrong; \paradoxbad{world prior replaces source}. \\
% \textbf{\textit{Overwrite}} & \textit{Recon.} Einstein was born in \paradoxbadhl{Germany}. \quad \textit{A:}  \paradoxbadhl{Germany} \paradoxwrong; \paradoxbad{world prior replaces source}. \\
% \end{tabular}
% \end{paradoxcasebox}

\textbf{Knowledge Overwriting  Evaluation.}
To isolate this phenomenon, we construct a controlled counterfactual evaluation based on FaithEval~\cite{ming2024faitheval} and ConflictQA~\cite{Xie2024KnowledgeConflict}. Both datasets introduce explicit factual contradictions into natural contexts (e.g., replacing ``Einstein was born in Germany'' with ``Einstein was born in France''). We retain unambiguous examples with established facts, ensuring a clear conflict between source evidence and parametric knowledge.

Each counterfactual context is compressed and paired with a question targeting the modified fact. Candidate answers include \textit{i) the source-consistent counterfactual answer} and \textit{ii) plausible world-knowledge alternatives}. We measure \textit{knowledge faithfulness} as the accuracy of selecting the source-consistent answer, i.e., prioritizing compressed evidence over internal priors. 
Dataset construction and filtering procedures are detailed are provided in Appendix~\ref{app:prompts}, with an optional candidate-generation prompt included for reference.
% Construction and filtering details are in Appendix~\ref{app:prompts}, with an optional candidate-generation prompt included for reference.
% Each counterfactual context is compressed and paired with a question targeting the modified fact. Candidate answers include \textit{i) the source-consistent counterfactual answer} and \textit{ii) plausible world-knowledge alternatives}. \textit{Knowledge faithfulness} is the accuracy of selecting the source-consistent answer. Construction and filtering details are in Appendix~\ref{app:prompts}.

% \begin{wrapfigure}{r}{0.49\textwidth}
% \vspace{-0.18in}
% \begin{paradoxcasebox}
% \tiny
% \setlength{\tabcolsep}{2pt}
% \renewcommand{\arraystretch}{0.94}
% \begin{tabular}{@{}p{0.24\linewidth}p{0.7\linewidth}@{}}
% \textbf{Input} & Einstein was born in \paradoxhl{France}. \textbf{Question:} birthplace? \\
% \textbf{Source-faithful} & Answer = \paradoxhl{France}; source evidence is retained. \paradoxok \\
% \textbf{Prior-overwrite} & Answer = \paradoxbadhl{Germany}; \paradoxbad{world prior overwrites source}. \\
% \end{tabular}
% \end{paradoxcasebox}
% \vspace{-0.18in}
% \end{wrapfigure}

\textbf{Results and Analysis.}
% Table~\ref{tab:accuracy} reveals a non-monotonic scale effect: increasing compressor size does not necessarily lead to higher faithfulness, and larger models are more susceptible to parametric-prior intrusion. 
Table~\ref{tab:accuracy} reveals a non-monotonic scale effect: larger compressors do not necessarily improve faithfulness and are more susceptible to parametric-prior intrusion. 
For example, at $64\times$, LLaMA-90B improves BLEU over LLaMA-3B (0.33 vs. 0.29) but drops clearly in overwriting accuracy (0.53 vs. 0.62). Notably, compressors around the 3--4B scale emerge as a particularly stable and effective regime, often achieving better faithfulness across model families.

\vskip -0.8in
\subsection{Dissection 2: Semantic Drift}
\label{subsec:drift}
\textit{Semantic drift} refers to fluent, topically coherent compression that nevertheless distorts the semantic or relational structure of the source.
Unlike knowledge overwriting, it arises from structural degradation during paraphrastic compression rather than conflict with parametric knowledge, with the case below highlighting a typical scope shift.

% \textit{Semantic drift} refers to a failure mode where compressor remains surface fluency and topical coherence while subtly distorting the semantic or relational structure of the source. Unlike knowledge overwriting, it arises from structural degradation during paraphrastic compression rather than conflict with parametric knowledge.
% rather than factual conflict

\begin{paradoxcasebox}
\small
\setlength{\tabcolsep}{2pt}
\renewcommand{\arraystretch}{0.80}
\begin{tabular}{@{}p{0.126\linewidth}@{\hspace{3pt}}p{0.9\linewidth}@{}}
\textbf{Input} & The \paradoxsrcred{dwarves} are funny, except maybe \paradoxsrcgreen{Thorin}. \quad \textit{Q:} exception? \\
\textbf{No drift} & \textit{Recons.} The dwarves are funny, though \paradoxreconhl{Thorin} may be an exception. \quad \textit{A:} Thorin \paradoxok \\
\textbf{Scope-drift} & \textit{Recons.} Except for \paradoxdrift{the dwarves}, Thorin is funny. \textit{A:} the dwarves \paradoxwrong; \paradoxbad{exception scope shifts} \\
\end{tabular}
\end{paradoxcasebox}

% \begin{paradoxcasebox}
% % \tiny
% % \scriptsize
% % \footnotesize
% \small
% \setlength{\tabcolsep}{2pt}
% \renewcommand{\arraystretch}{0.8}
% \begin{tabular}{@{}p{0.12\linewidth}p{0.90\linewidth}@{}}
% \textbf{\textit{Input}} & The \paradoxhl{dwarves} are funny, except maybe \paradoxhl{Thorin}. \qquad \qquad \textit{Q:} exception? \\
% \textbf{\textit{No drift}} & \textit{Recon.} The dwarves are funny, though \paradoxhl{Thorin} may be an exception. \quad \textit{A:} \paradoxhl{Thorin} \paradoxok  \\
% \textbf{\textit{Scope-drift}} & \textit{Recon.} Except for \paradoxbadhl{the dwarves}, Thorin is funny. \quad \textit{A:} \paradoxbadhl{the dwarves} \paradoxwrong; \paradoxbad{exception scope shifts}. \\
% \end{tabular}
% \end{paradoxcasebox}

\textbf{Semantic Drift Evaluation.}
% We evaluate seven common dimensions of semantic drift: \textit{main topic}, \textit{entity list}, \textit{predicate exactness}, \textit{relation anchor}, \textit{coreference}, \textit{role binding}, and \textit{modifier scope}. Together, they probe whether compression preserves the central subject, specific entities, predicates, discourse relations, pronoun links, role assignments, and qualifying constraints.
Existing benchmarks focus on topical relevance or factual recall, and are largely insensitive to structural distortions such as role reversal or modifier loss. We therefore build a diagnostic QA dataset over FineWeb~\cite{penedo2024fineweb} and FaithEval~\cite{ming2024faitheval}, using DeepSeek-R1~\cite{guo2025deepseek} to generate questions that require precise relations rather than topical gist.

% \begin{wrapfigure}{r}{0.52\textwidth}
% \vspace{-0.18in}
% \begin{paradoxcasebox}
% \tiny
% \setlength{\tabcolsep}{2pt}
% \renewcommand{\arraystretch}{0.94}
% \begin{tabular}{@{}p{0.20\linewidth}p{0.8\linewidth}@{}}
% \textbf{Input} & The \paradoxhl{dwarves} are funny, except maybe \paradoxhl{Thorin}. \textbf{Question:} exception? \\
% \textbf{Source-faithful} & The \paradoxhl{dwarves} are funny, except maybe \paradoxhl{Thorin}. \paradoxok \\
% \textbf{Scope-drift} & Except for \paradoxbadhl{the dwarves}, Thorin is funny; \paradoxbad{exception scope shifts}. \\
% \end{tabular}
% \end{paradoxcasebox}
% \vspace{-0.18in}
% \end{wrapfigure}

% We cover seven drift dimensions: \textit{main topic}, \textit{entity list}, \textit{predicate exactness}, \textit{relation anchor}, \textit{coreference}, \textit{role binding}, and \textit{modifier scope}, probing shifts in subjects, entities, predicates, discourse links, antecedents, roles, and qualifiers. Answerable questions require exact-substring answers from the source, and each context also includes plausible unanswerable questions to detect hallucinated details. Tab.~\ref{tab:callout_sd} illustrates how reconstruction can shift scope despite topical fluency. Tab.~\ref{tab:accuracy} reports the micro-averaged QA accuracy across all seven dimensions. Per-dimension results and generation/filtering details are in Appendices~\ref{app:7_dims}, \ref{app:prompt_semantic}, and \ref{app:qa_filtering}.

We cover seven drift dimensions: \textit{main topic}, \textit{entity list}, \textit{predicate exactness}, \textit{relation anchor}, \textit{coreference}, \textit{role binding}, and \textit{modifier scope}, probing shifts in subjects, entities, predicates, discourse links, antecedents, roles, and qualifiers. Answerable questions require exact-substring answers from the source; the callout below illustrates how reconstruction can shift scope despite topical fluency. Tab.~\ref{tab:accuracy} reports the micro-averaged QA accuracy across all seven dimensions. Per-dimension results and generation/filtering details are in Appendices~\ref{app:7_dims}, \ref{app:prompt_semantic}, and \ref{app:qa_filtering}.

% Specifically, we evaluate seven dimensions along which semantic drift frequently occurs: 
% (1) \textit{main topic}, testing whether the central subject or domain shifts; 
% (2) \textit{entity list}, detecting degradation of specific entities into coarse categories (e.g., ``Nike and Adidas'' $\rightarrow$ ``sportswear brands''); 
% (3) \textit{predicate exactness}, assessing preservation of key verbs and relational predicates; 
% (4) \textit{relation anchor}, verifying maintenance of causal, contrastive, or conditional links across sentences; 
% (5) \textit{coreference}, checking correct resolution of pronouns to their antecedents; 
% (6) \textit{role binding}, ensuring accurate ``who did what to whom'' assignments; and 
% (7) \textit{modifier scope}, testing retention of qualifiers and constraints that prevent over-generalization.

% For precision, answerable questions require exact substring answers from the original context, while each context also includes plausible but unanswerable questions to detect hallucinated details. 
% The main table reports the micro-averaged QA accuracy across all seven dimensions. We provide per-dimension results in Appendix~\ref{app:7_dims}, and the QA generation prompts and filtering criteria in Appendix~\ref{app:prompt_semantic}$\&$~\ref{app:qa_filtering}.

% The main table reports micro-averaged QA accuracy across all seven dimensions; per-dimension results and generation/filtering details are provided in Appendix~\ref{app:7_dims}, Appendix~\ref{app:prompt_semantic}, and Appendix~\ref{app:qa_filtering}.

\textbf{Results and Analysis.}
Tab.~\ref{tab:accuracy} shows the same non-monotonic scale effect for structural QA accuracy: fidelity improves up to the 3--4B range but declines for larger compressors, even as BLEU continues to increase. 
This divergence points to semantic drift rather than factual overwriting: larger models better preserve fluent semantic gist, but become less sensitive to fine-grained relational constraints. Overall, scaling appears to favor abstract semantic compression over precise structural preservation.
% This divergence reflects a failure mode distinct from factual overwriting. Larger models increasingly capture semantic gist and generate fluent reconstructions, yet become less sensitive to fine-grained relational constraints, leading to systematic semantic drift. Overall, scaling appears to favor abstract semantic compression at the expense of precise structural preservation.

% \subsection{Summary}
% The results in this section establish the \textbf{\textit{Size-Fidelity Paradox}} as a robust phenomenon. While scaling laws drive down training loss, they inadvertently amplify the model's tendency to overwrite facts and drift semantically. This suggests that existing evaluations relying on n-gram overlap (BLEU/ROUGE-L) are insufficient for context compression. They mask the critical trade-off between a model's generative capability and its reconstructive faithfulness.

%% file: tab/dataset.tex
\begin{wraptable}{r}{0.4\textwidth}
% \small
\footnotesize
% \scriptsize
% \tiny
\renewcommand{\arraystretch}{1.2} % 行高稍微加大
\setlength{\tabcolsep}{1.0pt}        % 适中列间距
\centering 
\vskip -0.26in
\caption{Task and dataset statistics.}
\vskip 0.04in
\label{tab:dataset}
\resizebox{\linewidth}{!}{%
\begin{tabular}{@{}c*{3}{c}@{}}
\toprule
Task & Dataset & \#Contexts & \#QAs \\
\midrule
Reconstruction & Fineweb & 4,444,672 & --\\
\midrule
\multirow{2}{*}{Knowledge Overwriting} & FaithEval & 598 & 598 \\
& ConflictQA & 1356 & 1356 \\
\midrule
\multirow{2}{*}{Semantic Drift} & Fineweb & 449 & 8980 \\
& FaithEval & 598 & 11960 \\

\bottomrule
\end{tabular}
}
\vskip -0.2in
\end{wraptable}

%% file: tab/metric.tex
\begin{wraptable}{r}{0.43\textwidth}
\scriptsize
\renewcommand{\arraystretch}{1.02}
\setlength{\tabcolsep}{3pt}
\centering
\vspace{-0.26in}
\caption{Standard reconstruction metrics for Qwen3 compressors at $4\times$ compression.}
\label{tab:metric}
\vspace{0.05in}
\resizebox{\linewidth}{!}{%
\begin{tabular}{@{}cccccc@{}}
\toprule
\#Params & BLEU & ROUGE-L & chrF++ & BERTScore & Avg. \\
\midrule
0.6B & 0.952 & 0.969 & 0.983 & 0.997 & 0.975 \\
1.7B & 0.941 & 0.957 & 0.975 & 0.995 & 0.967 \\
4B & 0.943 & 0.964 & 0.978 & 0.996 & 0.970 \\
8B & 0.940 & 0.955 & 0.969 & 0.995 & 0.965 \\
32B & \textbf{0.968} & \textbf{0.982} & \textbf{0.991} & \textbf{0.998} & \textbf{0.985} \\
\bottomrule
\end{tabular}%
}
\vspace{-0.12in}
\end{wraptable}

%% file: tab/accu_all_v3.tex
% \caption{\textbf{Quantitative Evaluation of the Size-Fidelity Paradox across Qwen and LLaMA Families (0.6B--90B).}
% We report BLEU as the representation of surface reconstruction metrics and two diagnostic QA evaluations under varying compression rates.
% \textbf{Bold} indicate the best model in the same family and compression group.
% shaded Avg./\textcolor{academicblue}{$\Delta$} columns denote category averages and drops from the group best.\textcolor{academicred}{red} marks the worst QA Avg./drop.}

\begin{table*}[!htbp]
\scriptsize
\renewcommand{\arraystretch}{0.90}
\setlength{\tabcolsep}{3.0pt}
% \definecolor{accuavgshade}{RGB}{242,242,242}
\definecolor{accuavgshade}{RGB}{240,243,248}
% \definecolor{accuavgshade}{RGB}{238,241,250}
\definecolor{academicblue}{RGB}{0,92,175}
\definecolor{academicred}{RGB}{180,35,35}
\providecommand{\bestdelta}{\textemdash\textemdash}
\providecommand{\accuavgcell}[1]{\cellcolor{accuavgshade}{#1}}
\providecommand{\accuavgredcell}[1]{\cellcolor{accuavgshade}{\textcolor{academicred}{#1}}}
\providecommand{\accudeltacell}[1]{\cellcolor{accuavgshade}{\textcolor{academicblue}{#1}}}
\providecommand{\accudeltaredcell}[1]{\cellcolor{accuavgshade}{\textcolor{academicred}{#1}}}
\providecommand{\accubestdeltacell}{\cellcolor{accuavgshade}{\bestdelta}}

\centering
\vskip -0.4in
\caption{\textbf{Quantitative Evaluation of the Size-Fidelity Paradox across Qwen and LLaMA Families (0.6B--90B).}
BLEU represents surface reconstruction, while diagnostic QA scores measure \textit{knowledge overwriting} and \textit{semantic drift} across compression rates.
\textbf{Bold} marks the best model within each family/compression group.
Shaded Avg./\textcolor{academicblue}{$\Delta$} columns denote category averages and drops from the group best; \textcolor{academicred}{red} marks the worst QA average/drop.
Full results are provided in Appendix~\ref{app:full_results}.}
\label{tab:accuracy}
\vskip 0.04in

\resizebox{0.99\textwidth}{!}{%
\begin{tabular}{@{}c c c c c c c c c c c c c@{}}
\toprule
\multirow{2}{*}{\raisebox{-0.7\height}{\makecell{Compression\\Rate}}}
& \multirow{2}{*}{Model}
& \multirow{2}{*}{\centering \#Params}
% & \multirow{2}{*}{\multicolumn{2}{c}{Reconstruction}}
& \multicolumn{2}{c}{\multirow{2}{*}{Reconstruction}}
% & \multicolumn{2}{c}{Reconstruction~(BLEU)}
& \multicolumn{4}{c}{\textit{Knowledge Overwriting}}
& \multicolumn{4}{c}{\textit{Semantic Drift}} \\
\cmidrule(r){6-9}\cmidrule(l){10-13}
% & & & 
& & & \multicolumn{2}{c}{}
% \makecell{FineWeb}
% & \accudeltacell{$\Delta$}
& FaithEval & ConflictQA & \accuavgcell{Avg.} & \accudeltacell{$\Delta$}
& FineWeb & FaithEval & \accuavgcell{Avg.} & \accudeltacell{$\Delta$} \\
\midrule

\multirow{9}{*}{4$\times$} & \multirow{5}{*}{Qwen 3} & 0.6b
& 0.95 & \accudeltacell{(-2.06\%)}
& 0.71 & \textbf{0.95} & \accuavgcell{0.83} & \accudeltacell{(-1.19\%)}
& 0.83 & \textbf{0.83} & \accuavgcell{\textbf{0.83}} & \accubestdeltacell \\
& & 1.7b
& 0.94 & \accudeltacell{(-3.09\%)}
& \textbf{0.76} & 0.89 & \accuavgcell{0.83} & \accudeltacell{(-1.19\%)}
& 0.83 & 0.81 & \accuavgcell{0.82} & \accudeltacell{(-1.20\%)} \\
& & 4b
& 0.94 & \accudeltacell{(-3.09\%)}
& 0.72 & \textbf{0.95} & \accuavgcell{\textbf{0.84}} & \accubestdeltacell
& \textbf{0.87} & 0.79 & \accuavgcell{\textbf{0.83}} & \accubestdeltacell \\
& & 8b
& 0.94 & \accudeltacell{(-3.09\%)}
& 0.74 & 0.91 & \accuavgcell{0.83} & \accudeltacell{(-1.19\%)}
& 0.79 & \textbf{0.83} & \accuavgcell{0.81} & \accudeltacell{(-2.41\%)} \\
& & 32b
& \textbf{0.97} & \accubestdeltacell
& 0.68 & 0.80 & \accuavgredcell{0.74} & \accudeltaredcell{(-11.90\%)}
& 0.60 & 0.78 & \accuavgredcell{0.69} & \accudeltaredcell{(-16.87\%)} \\
\cmidrule{2-13}
& \multirow{4}{*}{LLaMA 3.2} & 1b
& 0.94 & \accudeltacell{(-3.09\%)}
& \textbf{0.77} & 0.88 & \accuavgcell{\textbf{0.83}} & \accubestdeltacell
& 0.78 & \textbf{0.85} & \accuavgcell{\textbf{0.82}} & \accubestdeltacell \\
& & 3b
& 0.90 & \accudeltacell{(-7.22\%)}
& 0.74 & \textbf{0.91} & \accuavgcell{\textbf{0.83}} & \accubestdeltacell
& \textbf{0.79} & 0.84 & \accuavgcell{\textbf{0.82}} & \accubestdeltacell \\
& & 11b
& 0.92 & \accudeltacell{(-5.15\%)}
& 0.65 & 0.80 & \accuavgcell{0.73} & \accudeltacell{(-12.05\%)}
& 0.64 & 0.76 & \accuavgcell{0.70} & \accudeltacell{(-14.63\%)} \\
& & 90b
& \textbf{0.97} & \accubestdeltacell
& 0.60 & 0.81 & \accuavgredcell{0.71} & \accudeltaredcell{(-14.46\%)}
& 0.59 & 0.68 & \accuavgredcell{0.64} & \accudeltaredcell{(-21.95\%)} \\

\cmidrule{1-13}

\multirow{9}{*}{16$\times$} & \multirow{5}{*}{Qwen 3} & 0.6b
& 0.83 & \accudeltacell{(-3.49\%)}
& 0.70 & \textbf{0.94} & \accuavgcell{0.82} & \accudeltacell{(-1.20\%)}
& 0.75 & 0.84 & \accuavgcell{0.80} & \accudeltacell{(-3.61\%)} \\
& & 1.7b
& \textbf{0.86} & \accubestdeltacell
& \textbf{0.75} & 0.88 & \accuavgcell{0.82} & \accudeltacell{(-1.20\%)}
& 0.79 & \textbf{0.86} & \accuavgcell{\textbf{0.83}} & \accubestdeltacell \\
& & 4b
& \textbf{0.86} & \accubestdeltacell
& 0.71 & \textbf{0.94} & \accuavgcell{\textbf{0.83}} & \accubestdeltacell
& \textbf{0.82} & 0.83 & \accuavgcell{\textbf{0.83}} & \accubestdeltacell \\
& & 8b
& 0.84 & \accudeltacell{(-2.33\%)}
& 0.72 & 0.89 & \accuavgcell{0.81} & \accudeltacell{(-2.41\%)}
& 0.73 & 0.79 & \accuavgcell{0.76} & \accudeltacell{(-8.43\%)} \\
& & 32b
& 0.85 & \accudeltacell{(-1.16\%)}
& 0.61 & 0.72 & \accuavgredcell{0.67} & \accudeltaredcell{(-19.28\%)}
& 0.57 & 0.74 & \accuavgredcell{0.66} & \accudeltaredcell{(-20.48\%)} \\
\cmidrule{2-13}
& \multirow{4}{*}{LLaMA 3.2} & 1b
& 0.82 & \accudeltacell{(-4.65\%)}
& 0.72 & 0.84 & \accuavgcell{0.78} & \accudeltacell{(-4.88\%)}
& \textbf{0.75} & \textbf{0.90} & \accuavgcell{\textbf{0.83}} & \accubestdeltacell \\
& & 3b
& 0.81 & \accudeltacell{(-5.81\%)}
& \textbf{0.73} & \textbf{0.90} & \accuavgcell{\textbf{0.82}} & \accubestdeltacell
& \textbf{0.75} & \textbf{0.90} & \accuavgcell{\textbf{0.83}} & \accubestdeltacell \\
& & 11b
& 0.85 & \accudeltacell{(-1.16\%)}
& 0.63 & 0.77 & \accuavgcell{0.70} & \accudeltacell{(-14.63\%)}
& 0.61 & 0.73 & \accuavgcell{0.67} & \accudeltacell{(-19.28\%)} \\
& & 90b
& \textbf{0.86} & \accubestdeltacell
& 0.55 & 0.74 & \accuavgredcell{0.65} & \accudeltaredcell{(-20.73\%)}
& 0.56 & 0.63 & \accuavgredcell{0.60} & \accudeltaredcell{(-27.71\%)} \\

\cmidrule{1-13}

\multirow{9}{*}{64$\times$} & \multirow{5}{*}{Qwen 3} & 0.6b
& 0.17 & \accudeltacell{(-41.38\%)}
& 0.44 & 0.59 & \accuavgredcell{0.52} & \accudeltaredcell{(-17.74\%)}
& 0.31 & 0.39 & \accuavgcell{0.35} & \accudeltacell{(-20.45\%)} \\
& & 1.7b
& 0.21 & \accudeltacell{(-27.59\%)}
& 0.51 & 0.60 & \accuavgcell{0.56} & \accudeltacell{(-11.29\%)}
& 0.35 & 0.42 & \accuavgcell{0.39} & \accudeltacell{(-11.36\%)} \\
& & 4b
& 0.27 & \accudeltacell{(-6.90\%)}
& \textbf{0.54} & \textbf{0.70} & \accuavgcell{\textbf{0.62}} & \accubestdeltacell
& \textbf{0.41} & 0.46 & \accuavgcell{\textbf{0.44}} & \accubestdeltacell \\
& & 8b
& 0.28 & \accudeltacell{(-3.45\%)}
& \textbf{0.54} & 0.67 & \accuavgcell{0.61} & \accudeltacell{(-3.23\%)}
& 0.38 & \textbf{0.49} & \accuavgcell{\textbf{0.44}} & \accubestdeltacell \\
& & 32b
& \textbf{0.29} & \accubestdeltacell
& 0.47 & 0.56 & \accuavgredcell{0.52} & \accudeltaredcell{(-17.74\%)}
& 0.27 & 0.35 & \accuavgredcell{0.31} & \accudeltaredcell{(-29.55\%)} \\
\cmidrule{2-13}
& \multirow{4}{*}{LLaMA 3.2} & 1b
& 0.19 & \accudeltacell{(-42.42\%)}
& 0.49 & 0.57 & \accuavgredcell{0.53} & \accudeltaredcell{(-14.52\%)}
& 0.33 & 0.44 & \accuavgcell{0.39} & \accudeltacell{(-7.32\%)} \\
& & 3b
& 0.29 & \accudeltacell{(-12.12\%)}
& \textbf{0.55} & \textbf{0.68} & \accuavgcell{\textbf{0.62}} & \accubestdeltacell
& \textbf{0.35} & \textbf{0.47} & \accuavgcell{\textbf{0.41}} & \accubestdeltacell \\
& & 11b
& 0.30 & \accudeltacell{(-9.09\%)}
& 0.51 & 0.63 & \accuavgcell{0.57} & \accudeltacell{(-8.06\%)}
& 0.24 & 0.33 & \accuavgcell{0.29} & \accudeltacell{(-31.71\%)} \\
& & 90b
& \textbf{0.33} & \accubestdeltacell
& 0.45 & 0.61 & \accuavgredcell{0.53} & \accudeltaredcell{(-14.52\%)}
& 0.24 & 0.31 & \accuavgredcell{0.28} & \accudeltaredcell{(-34.15\%)} \\

\bottomrule
\end{tabular}%
}

\vskip -0.2in
\end{table*}

%% file: 5_generation.tex
\vskip -0.2in
\section{Generalization and Downstream Impact}

\subsection{Generalization Analysis}
\label{sec:generalization}

% To verify that the Size--Fidelity Paradox does not arise from a particular experimental configuration, we examine the transferability of our findings under different generalization conditions. We vary the decoder scale and family, replace the pretraining corpus with a scientific-domain corpus, and sweep the weights of reconstruction and next-token prediction losses to account for possible scale-dependent optimization patterns. Unless otherwise specified, all experiments follow the same compression setting as the main study and vary only the factor under analysis.

To verify that the Size--Fidelity Paradox does not arise from a particular experimental configuration, we examine its transferability through \textbf{\textit{(1) decoder compatibility}}, \textbf{\textit{(2) cross-domain robustness}}, and \textbf{\textit{(3) loss-weight sensitivity experiments}}. These settings vary the decoder, the pretraining corpus, and the training-objective balance, respectively, while keeping the main compression protocol unchanged.

\subsubsection{Decoder Compatibility}
In our main experiments, we use a fixed Meta-LLaMA-3.2-8B-Instruct decoder. In this section, we train compressors from the Qwen3 family (0.6B, 4B, 8B) at 16× compression and evaluate them with Qwen3 decoders of varying capacity: 0.6B and 4B. This cross-family transfer (LLaMA→Qwen) tests whether the paradox generalizes beyond architectural specifics, while the decoder size ablation tests whether smaller decoders expose or suppress the fidelity degradation we observe in larger models.

\input{tab/generalization_1_decoder}

\textbf{Robustness to Decoder Scaling.}
Tab.~\ref{tab:generalization_1_decoder} presents reconstruction accuracy and QA performance across decoder configurations. When the decoder size is reduced from 8B to 4B, we observe a slight baseline drop in reconstruction metrics. However, the paradox remains pronounced: the 8B compressor continues to exhibit significantly higher rates of knowledge overwriting and semantic drift compared to the 4B compressor.

When using the 0.6B decoder, all compressors experience substantial degradation in both reconstruction and QA tasks, reflecting the decoder's limited generation capacity. Crucially, this bottleneck narrows the performance gap between compressors—the 8B compressor's tendency toward overwriting and drift is partially suppressed by the decoder's constraints. However, the relative ordering persists: even under a weak decoder, the 8B compressor shows higher error rates on both QA tasks compared to the 0.6B and 4B compressors. This suggests that while decoder capacity modulates the severity of the paradox, it does not eliminate the underlying mechanism. 
We further pair Qwen3-32B with stronger decoders, including a matched 32B decoder, in Appendix~\ref{app:large_decoder}; it still underperforms the 4B baseline on fidelity-oriented QA, suggesting that decoder under-capacity is not the main cause.

% \textbf{Universality Across Families.}
% Crucially, Tab.~\ref{tab:generalization_1_decoder} involves pairing Qwen-family compressors with Qwen-family decoders, whereas our main experiments used LLaMA-family decoders. The persistence of the paradox across this architectural transfer confirms that fidelity loss stems from scaled compressor representations, rather than a compatibility artifact between specific model families.

\subsubsection{Cross-domain Robustness}
\input{tab/generalization_2_cross_domain}

To assess the out-of-domain generalizability of the paradox, we additionally pretrain Qwen3 compressors~(at 16$\times$ compression) on ArxivCorpus~($\approx$20.9B tokens)~\cite{clement2019arxiv}, a scientific corpus of academic abstracts, and apply the same reconstruction and diagnostic QA protocols. As shown in Tab.~\ref{tab:generalization_2_cross_domain}, reconstruction improves with model scale, while fidelity-oriented QA remains non-monotonic: the larger compressor underperforms the mid-sized compressor on both knowledge-overwriting and semantic-drift probes. This confirms that the scale--fidelity tradeoff persists beyond the original FineWeb pretraining domain.

\subsubsection{Loss-weight Sensitivity}
\label{sec:loss_sweep}
\input{tab/generalization_3_loss_weight}

A possible concern is that different model scales may require different balances between reconstruction and next-token prediction during pretraining. 
We therefore conduct a $\lambda$-sweep, where $\lambda$ weights the next-token prediction loss$\mathcal{L}_{\rm nt}$ and $(1-\lambda)$ controls the reconstruction loss$\mathcal{L}_{\rm re}$,
% \begin{equation}
%     \mathcal{L}=(1-\lambda)\mathcal{L}_{\rm re}+\lambda\mathcal{L}_{\rm nt},
% \end{equation}
and evaluate Qwen3 compressors~(at 4$\times$ compression) with the same reconstruction and diagnostic QA protocols.

Tab.~\ref{tab:generalization_3_loss_weight} shows that the loss balance shifts the reconstruction--fidelity tradeoff, but does not restore monotonic scaling in fidelity-oriented QA. Larger compressors still fail to consistently outperform smaller or mid-sized ones on knowledge overwriting and semantic drift, suggesting that the scale--fidelity tradeoff is not merely an artifact of applying the same loss mixture to models of different sizes.

% \subsection{Downstream Impact on Fidelity-Sensitive Tasks}
\subsection{Downstream Impact on RAG-based QA and ICL}
\label{sec:downstream}

To further examine downstream effects of the scale--fidelity tradeoff, we evaluate Qwen3 compressors from 0.6B to 32B on retrieval-augmented generative question answering~(RAG-based QA) and in-context learning~(ICL).

\input{tab/downstream_1}

For RAG-based QA, each example is a ({context}, {question}, {answer}) triple. We compare three settings: (i)~\textit{w/o Context}, where the inference model receives only the question; (ii) \textit{w/ Context}, where it receives the original context; and (iii) \textit{4$\times$ Compressed Context}, where the context is compressed at 4$\times$ and reconstructed by the fixed decoder before QA. We evaluate on MuSiQue~\cite{trivedi2021musique}, which requires connected multi-hop evidence composition, and QASPER~\cite{Dasigi2021ADO}, which targets information-seeking QA over scientific documents, using F1 and Exact Match~(EM).

For ICL, we evaluate five BBH subsets~\cite{suzgun2022challenging, li2024dialogue, li2024concentrate}: \textit{Date Understanding}, \textit{Causal Judgement}, \textit{Logical Deduction}, \textit{Formal Fallacies}, and \textit{Tracking Shuffled Objects}, reporting EM accuracy over 250/187 evaluation examples for each subset. We compare (i)~\textit{Zero-shot}, (ii)~\textit{Full ICL}, and (iii)~\textit{4$\times$ Compressed ICL}; in the compressed setting, only the demonstration block is compressed and reconstructed, while the test query remains unchanged. 
% Detailed protocols are provided in \red{Appendix}~\ref{app:downstream_details}.

Tab.~\ref{tab:downstream_1} reports the results on RAG-based QA. Across MuSiQue and QASPER, the 8B compressor performs worse than the 0.6B and 1.7B compressors, with a similar decline observed for 32B. 

\input{tab/downstream_2}
Tab.~\ref{tab:downstream_2} reports the results on BBH ICL tasks, where compressing the demonstrations with larger compressors again leads to lower accuracy, especially for the 8B and 32B models. 
% These results show that the Size--Fidelity Paradox can extend to downstream tasks whose predictions depend on source-specific facts, relations, or task demonstrations. They suggest that compressor scale should be chosen carefully when downstream performance relies on faithful preservation of the provided context.
Overall, the downstream results align with our diagnostic findings: when task performance depends on source-specific facts, relations, or demonstrations, larger compressors can be less reliable despite stronger reconstruction signals. This suggests that compressor scale should be considered together with the fidelity requirements of the target application.

%% file: tab/generalization_1_decoder.tex
\begin{wraptable}{l}{0.52\textwidth}
% \small
\footnotesize
% \scriptsize
% \tiny
\renewcommand{\arraystretch}{1.0} % 行高稍微加大
\setlength{\tabcolsep}{1.0pt}        % 适中列间距
\centering 
\vskip -0.26in
% \caption{Decoder generalizability ablation. Quantitative evaluation of Qwen3 compressors (at 16$\times$ compression) utilizing Qwen3-0.6B and Qwen3-4B as decoders. }
\caption{Decoder generalizability ablation with Qwen3 compressors at 16$\times$ compression and Qwen3-0.6B/4B decoders.}
\label{tab:generalization_1_decoder}
\vskip 0.08in
\resizebox{\linewidth}{!}{%
\begin{tabular}{@{}c*{5}{c}@{}}
\toprule
\multirow{2}{*}{\raisebox{-0.7\height}{\makecell{Compressor}}} 
& \multicolumn{1}{c}{Recons.$\uparrow$}& \multicolumn{2}{c}{ QA(i)$\uparrow$ }  & \multicolumn{2}{c}{QA(ii)$\uparrow$ } \\

\cmidrule(r){2-2}\cmidrule(r){3-4}\cmidrule(r){5-6}
 & {\centering Fineweb} &{\centering FaithEval}  &{\centering ConflictQA} &  {\centering Fineweb} &  {\centering FaithEval}  \\
\midrule
\multicolumn{6}{c}{\textit{Decoder: Qwen3-0.6B \quad Compression Rate:16x}}\\
    \cmidrule{1-6}
                 Qwen3-0.6b & 0.62 & 0.59 & 0.63& 0.58& 0.66  \\
                 Qwen3-4b &0.66 & 0.54 & 0.62& 0.58& 0.68  \\
                 Qwen3-8b & 0.61 & 0.53 & 0.59& 0.52& 0.61  \\
    \cmidrule{1-6}
\multicolumn{6}{c}{\textit{Decoder: Qwen3-4B \quad Compression Rate:16x}}\\
    \cmidrule{1-6}
                 Qwen3-0.6b & 0.76 & 0.65 & 0.80& 0.62& 0.75  \\
                 Qwen3-4b & 0.79 & 0.66 & 0.81& 0.63& 0.77  \\
                 Qwen3-8b & 0.77 & 0.62 & 0.75 & 0.57& 0.71  \\
\bottomrule
\end{tabular}
}

\vskip -0.3in
\end{wraptable}

%% file: tab/generalization_2_cross_domain.tex
\begin{wraptable}{r}{0.52\textwidth}
\vspace{-0.12in}
\footnotesize
\renewcommand{\arraystretch}{1.0}
\setlength{\tabcolsep}{2.0pt}
\centering
\vskip -0.16in
\caption{Cross-domain robustness on ArxivCorpus.}
\label{tab:generalization_2_cross_domain}
\vskip 0.04in
\resizebox{\linewidth}{!}{%
\begin{tabular}{@{}lccccc@{}}
\toprule
\multirow{2}{*}{Compressor}
& \multicolumn{1}{c}{Recons.$\uparrow$}
& \multicolumn{2}{c}{QA(i)$\uparrow$}
& \multicolumn{2}{c}{QA(ii)$\uparrow$} \\
\cmidrule(lr){2-2}\cmidrule(lr){3-4}\cmidrule(lr){5-6}
& ArxivCorpus
& FaithEval & ConflictQA
& Fineweb & FaithEval \\
\midrule
Qwen3-0.6B & 0.72 & 0.60 & 0.77 & 0.71 & 0.76 \\
Qwen3-4B   & 0.76 & \textbf{0.69} & \textbf{0.82} & \textbf{0.79} & \textbf{0.81} \\
Qwen3-8B   & \textbf{0.77} & 0.63 & 0.79 & 0.74 & 0.77 \\
\bottomrule
\end{tabular}
}
\vskip -0.08in
\end{wraptable}

%% file: tab/generalization_3_loss_weight.tex
\begin{wraptable}{r}{0.52\textwidth}
\vspace{-0.12in}
\footnotesize
\renewcommand{\arraystretch}{1.0}
\setlength{\tabcolsep}{1.6pt}
\centering
\vskip -0.16in
\caption{Loss-weight sensitivity at 4$\times$ compression.}
\label{tab:generalization_3_loss_weight}
\vskip 0.04in
\resizebox{\linewidth}{!}{%
\begin{tabular}{@{}llccccc@{}}
\toprule
\multirow{2}{*}{Conpressor}
& \multirow{2}{*}{$\lambda$}
& \multicolumn{1}{c}{Recons.$\uparrow$}
& \multicolumn{2}{c}{QA(i)$\uparrow$}
& \multicolumn{2}{c}{QA(ii)$\uparrow$} \\
\cmidrule(lr){3-3}\cmidrule(lr){4-5}\cmidrule(lr){6-7}
& 
& FineWeb
& Faith. & Conf.
& FineWeb & Faith. \\
\midrule
\multirow{5}{*}{Qwen3-0.6B}
& 0.1 & 0.90 & 0.69 & 0.89 & 0.77 & 0.75 \\
& 0.3 & 0.93 & \textbf{0.71} & 0.92 & 0.80 & 0.79 \\
& 0.5 {\scriptsize(default)} & \textbf{0.95} & \textbf{0.71} & \textbf{0.95} & \textbf{0.83} & \textbf{0.83} \\
& 0.7 & 0.91 & 0.69 & 0.90 & 0.78 & 0.77 \\
& 0.9 & 0.87 & 0.66 & 0.85 & 0.74 & 0.71 \\
\midrule
\multirow{3}{*}{Qwen3-4B}
& 0.3 & \textbf{0.94} & 0.70 & 0.91 & 0.86 & 0.77 \\
& 0.5 {\scriptsize(default)} & \textbf{0.94} & \textbf{0.72} & \textbf{0.95} & \textbf{0.87} & \textbf{0.79} \\
& 0.7 & 0.91 & 0.68 & 0.90 & 0.82 & 0.78 \\
\bottomrule
\end{tabular}
}
\vskip -0.2in
\end{wraptable}

%% file: tab/downstream_1.tex
\begin{wraptable}{l}{0.44\textwidth}
\vspace{-0.27in}
% \small
\footnotesize
% \scriptsize
% \tiny
\renewcommand{\arraystretch}{1.0} % 行高稍微加大
\setlength{\tabcolsep}{1.0pt}        % 适中列间距
\centering 
\caption{Results on RAG-based QA.}
\label{tab:downstream_1}
\vspace{0.04in}
\resizebox{\linewidth}{!}{%
\begin{tabular}{@{}c*{5}{c}@{}}
\toprule
\multirow{2}{*}{\raisebox{-0.7\height}{\makecell{RAG-based QA}}} 
&  \multicolumn{2}{c}{ MuSiQue }  & \multicolumn{2}{c}{ QASPER } \\

\cmidrule(r){2-3}\cmidrule(r){4-5}
 & {\centering F1(\%)$\uparrow$} &{\centering EM(\%)$\uparrow$ }  & {\centering F1(\%)$\uparrow$} &{\centering EM(\%)$\uparrow$ } \\
\midrule
% \multicolumn{6}{c}{\textit{Decoder: Qwen3-0.6B \quad Compression Rate:16x}}\\
    \cmidrule{1-5}
                 w/o Context & 9.11 & 2.67 & 13.34 & 5.87  \\
                 w/ Context & \textbf{33.26} & \textbf{23.91} & \textbf{46.22} & \textbf{31.90}   \\
                 Qwen3-0.6b 4$\times$ & \underline{24.14} & \underline{15.08} & 28.93 & 19.16   \\
                 Qwen3-1.7b 4$\times$ & 22.67 & 13.65 & \underline{29.46} & \underline{19.37}   \\
                 Qwen3-4b 4$\times$ & 20.75 & 11.31 & 28.06 & 18.82   \\
                 Qwen3-8b 4$\times$ & 16.93 & 9.10 & 24.35 & 15.72   \\
                 Qwen3-32b 4$\times$ & 14.55 & 0.53 & 23.94 & 13.18   \\
\bottomrule
\end{tabular}
}

\vskip -0.2in
\end{wraptable}
% \vskip -0.15in

%% file: tab/downstream_2.tex
\begin{wraptable}{l}{0.44\textwidth}
\vskip -0.1in
% \small
\footnotesize
% \scriptsize
% \tiny
\renewcommand{\arraystretch}{1.0} % 行高稍微加大
\setlength{\tabcolsep}{1.0pt}        % 适中列间距
\centering 
% \caption{Downstream Impact on ICL}
\caption{Results on ICL.}
\label{tab:downstream_2}
\vspace{0.04in}
\resizebox{\linewidth}{!}{%
\begin{tabular}{@{}c*{6}{c}@{}}
\toprule
\multirow{2}{*}{\raisebox{-0.7\height}{\makecell{ICL}}} 
&  \multicolumn{6}{c}{ \centering BBH / EM(\%)$\uparrow$ } \\

\cmidrule(r){2-7}
 & Date. & Causal. & Logic. & Formal. & Track. & Avg. \\
 % & \#A & \#B & \#C & \#D & \#E & Avg. \\
\midrule
                 Zero-shot & 43.2 & 52.9 & 49.6 & 53.6 & 50.8 & 50.0 \\
                 Full ICL & \textbf{62.8} & \textbf{75.4} & \textbf{70.4} & \textbf{76.0} & \textbf{67.2} & \textbf{70.4} \\
                 Qwen3-0.6b 4$\times$ & 58.4 & \underline{70.6} & 66.8 & \underline{69.6} & \underline{60.4} & \underline{65.2} \\
                 Qwen3-1.7b 4$\times$ & \underline{58.8} & 70.1 & 66.8 & 69.2 & 59.2 & 64.8 \\
                 Qwen3-4b 4$\times$ & 57.2 & 69.0 & \underline{67.6} & 67.6 & 56.8 & 63.6 \\
                 Qwen3-8b 4$\times$ & 55.6 & 65.2 & 63.2 & 65.2 & 54.8 & 60.8\\
                 Qwen3-32b 4$\times$ & 55.2 & 64.2 & 63.6 & 64.0 & 57.6 & 60.9 \\
\bottomrule
\end{tabular}
}

\vskip -0.14in
\end{wraptable}

%% file: 6_analysis_v1.tex
\section{Beyond Scale: How Compression Changes with Model Size?}
\label{sec:analysis}

% Having established the Size-Fidelity Paradox, we now look beyond model size. Parameter count is only a coarse descriptor: fidelity in context compression depends on how source information is represented in the memory embeddings $\mathbf{Z} \in \mathbb{R}^{m \times d}$ (defined in Sec.~\ref{sec:principle}), and how tightly this representation constrains reconstruction. 
% We therefore probe $\mathbf{Z}$ from two complementary perspectives: memory organization, which we study via effective rank and relate to \textit{knowledge overwriting}; and reconstruction determinacy, which we study via conditional entropy and relate to \textit{semantic drift}.

Having established the Size-Fidelity Paradox, we now look beyond parameter count. Fidelity in context compression depends on what the memory embeddings $\mathbf{Z} \in \mathbb{R}^{m \times d}$ (defined in Sec.~\ref{sec:principle}) preserve about the source, and how tightly they constrain reconstruction.

We follow the same diagnostic logic for two probes: first identify how the probe changes with scale, then use fixed-scale training dynamics to interpret whether the compression objective favors high or low values, and finally test whether the probe predicts the failure mode implied by its meaning. Effective rank probes memory organization and \textit{knowledge overwriting}; conditional entropy probes reconstruction determinacy and \textit{semantic drift}.

% We therefore probe $\mathbf{Z}$ from two complementary perspectives: how source information is organized in memory, and how deterministically the memory supports source-specific reconstruction.

\subsection{How Scale Reorganizes Memory: Source Alignment vs. Semantic Dispersion}
\label{subsec:rank_analysis}

\begin{figure*}[ht]
  \vskip -0.12in
  \begin{center}
    \centerline{\includegraphics[width=0.9\columnwidth]{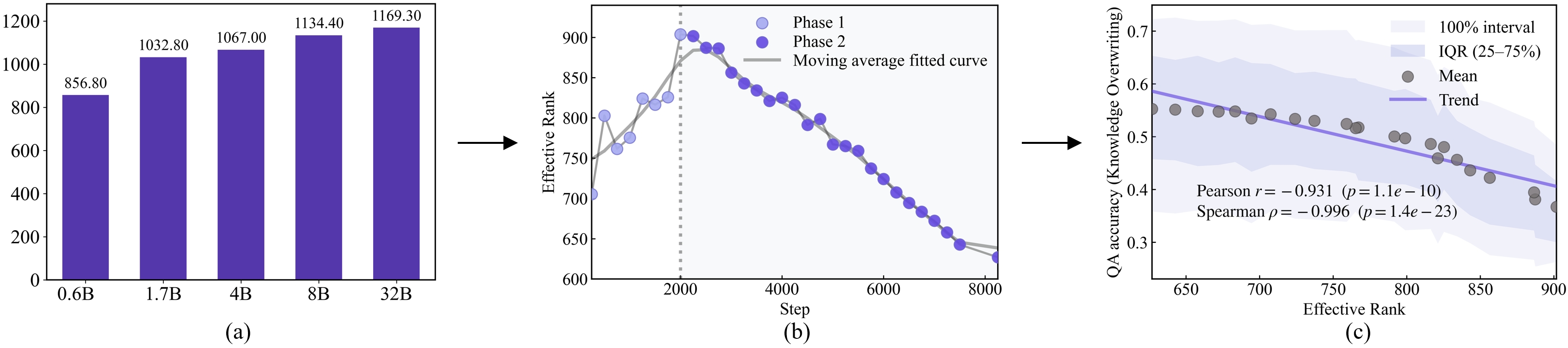}}
    \vskip -0.12in
    % \caption{(a) Effective rank increases monotonically with model scale in the Qwen3 family (0.6B–32B). (b) \textbf{Training dynamics of effective rank.} A clear two-phase trajectory emerges: early expansion followed by compression. (c) \textbf{Effective rank vs. QA performance.} Effective rank is negatively correlated with QA accuracy; shaded bands indicate the sample-level distribution.}
    \caption{\textbf{Effective-rank diagnosis.} (a) Scaling broadens the active memory subspace; (b) fixed-scale training compresses it after early expansion; (c) higher post-peak rank predicts lower knowledge-faithfulness QA accuracy.}
    \vskip -0.24in
    \label{fig:analysis_rank}
  \end{center}
\end{figure*}

% Knowledge overwriting suggests that larger compressors may not simply lose information at random. Instead, source-specific facts can be absorbed into broader semantic associations, allowing parametric priors to replace the evidence contained in the compressed context. We study this possibility by analyzing the geometry of the memory embeddings $\mathbf{Z}$.

\textit{Knowledge overwriting} is a source-anchoring failure: when the compressed context conflicts with parametric knowledge, reconstruction may recover a plausible prior fact rather than the source fact. If this failure is tied to memory geometry, the key question is whether the active subspace remains aligned with the particular input instead of expanding into nearby semantic alternatives.

We use effective rank as a probe of how broadly the compressed memory uses its latent space. Given the batch-stacked memory matrix $\mathbf{Z}_{\text{batch}}$, we compute
\begin{equation}
    \text{erank}(\mathbf{Z})
=
\exp\left(-\sum_{i=1}^{r} p_i \log p_i\right),
\qquad
p_i=\frac{\sigma_i}{\sum_{j=1}^{r}\sigma_j},
\end{equation}
where $\sigma_i$ are the singular values of $\mathbf{Z}_{\mathrm{batch}}$. Higher effective rank means that representation mass spreads across more active directions~(Appendix~\ref{app:erank_hidden} reports $\mathrm{erank}/d$ across scales). Such spread can encode richer semantic neighborhoods, but faithful compression requires source-aligned directions.

The scale sweep gives the first clue: effective rank increases with compressor size, so larger compressors activate a broader memory subspace (Fig.~\ref{fig:analysis_rank}(a)). This helps explain why scaling can improve surface reconstruction while weakening fidelity. A broader subspace may preserve the source gist, yet keep prior-compatible alternatives close enough to compete with rare or counterfactual facts.

Training dynamics reveal the sign of this probe. With model size fixed, rank first expands then contracts (Fig.~\ref{fig:analysis_rank}(b)): early optimization opens semantic directions to separate inputs, while later optimization prunes toward a source-sufficient structure. Thus, high rank is not inherently desirable; after basic reconstruction is learned, compression favors a compact, source-aligned subspace.

This interpretation predicts the failure link: if rank measures semantic breadth, excessive rank should hurt when source evidence must override prior knowledge. In the post-peak regime, higher effective rank correlates with lower knowledge-faithfulness QA accuracy (Fig.~\ref{fig:analysis_rank}(c)). 
The rank-intervention results in Appendix~\ref{app:rank_intervention} supports this structure: moderate spectral truncation improves fidelity by trimming the low-energy tail, whereas aggressive truncation removes source-bearing directions and hurts reconstruction. The useful bottleneck is not the smallest one, but one whose leading subspace preserves source evidence without inviting unnecessary variation.

\subsection{Memory Constraints: Source Recovery vs. Plausible Alternatives}
\label{subsec:entropy_analysis}

% Semantic drift reflects a different failure mode. The reconstruction may preserve topic and fluency, yet alter relations, roles, modifiers, or causal structure. This suggests that the compressed memory contains enough information for a plausible reconstruction, but not enough to force the exact source-specific one. We examine this issue through conditional entropy, measured under the same decoder conditioned on different compressor memories.

\textit{Semantic drift} is an under-constraint failure: reconstruction remains topical and fluent, but changes roles, modifiers, or causal relations. This suggests that memory supports plausible continuations without forcing the source-specific one, which conditional entropy probes under the same decoder.

% Given $\mathbf{Z}=f_\theta(\mathbf{x})$, the decoder defines
% $P_\phi(\mathbf{x}\mid \mathbf{Z}) = \prod_{t=1}^{L}P_\phi(x_t\mid x_{<t},\mathbf{Z})$.
% At each reconstruction step, we compute the teacher-forced token entropy
% $H_t(\mathbf{Z}) = -\sum_{v\in\mathcal{V}} P_\phi(v\mid x_{<t},\mathbf{Z}) \log P_\phi(v\mid x_{<t},\mathbf{Z})$,
% and average it over tokens to obtain $H(\mathbf{Z})$. Teacher forcing removes free-decoding variability, so this entropy measures how strongly memory constrains the source token path. Low entropy indicates determinate reconstruction; high entropy leaves multiple plausible continuations competitive.

Given $\mathbf{Z}=f_\theta(\mathbf{x})$, the decoder defines
$P_\phi(\mathbf{x}\mid \mathbf{Z}) = \prod_{t=1}^{L}P_\phi(x_t\mid x_{<t},\mathbf{Z})$.
At each reconstruction step, we compute the teacher-forced token entropy
$H_t(\mathbf{Z}) = -\sum_{v\in\mathcal{V}} P_\phi(v\mid x_{<t},\mathbf{Z}) \log P_\phi(v\mid x_{<t},\mathbf{Z})$,
and average it over tokens to obtain $H(\mathbf{Z})$. By conditioning on the gold prefix, this entropy abstracts away sampling-induced variation and directly reflects how tightly the memory constrains the next source token. Low entropy indicates determinate reconstruction, whereas high entropy means that multiple plausible continuations remain competitive.

% and average it over tokens to obtain $H(\mathbf{Z})$. Since this quantity is computed under teacher forcing, it removes variability from free-form decoding and measures how strongly the memory constrains the source token path. Low entropy indicates a determinate reconstruction; high entropy indicates that multiple locally plausible continuations remain competitive.

\begin{figure*}[ht]
  \vskip -0.12in
  \begin{center}
    \centerline{\includegraphics[width=0.9\columnwidth]{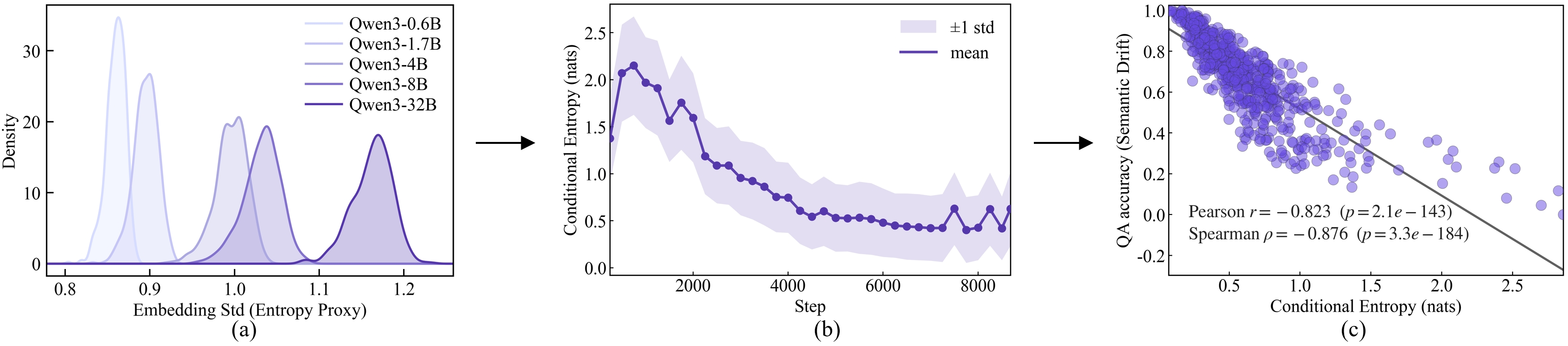}}
    \vskip -0.12in
    \caption{\textbf{Conditional-entropy diagnosis.} (a) Larger compressors yield higher conditional entropy, leading to less determinate reconstruction; (b) fixed-scale training reduces conditional entropy; (c) higher conditional entropy predicts lower semantic-drift QA accuracy.}
    \vskip -0.24in
    \label{fig:analysis_entropy}
  \end{center}
\end{figure*}

The scale trend identifies conditional entropy as the probe: larger compressors show higher entropy than mid-sized ones (Fig.~\ref{fig:analysis_entropy}(a); Appendix~\ref{app:entropy_theory}). Training again gives the sign. For a fixed model, entropy steadily decreases as optimization sharpens the source token path (Fig.~\ref{fig:analysis_entropy}(b)). Scaling therefore counteracts training, leaving more plausible alternatives under the same bottleneck.

This ambiguity is what semantic drift exposes. Higher conditional entropy correlates with lower semantic-drift QA accuracy (Fig.~\ref{fig:analysis_entropy}(c)): when memory does not sharply select the source continuation, role bindings can shift, modifiers can attach to the wrong entity, and causal relations can be smoothed into generic associations. The output remains coherent, but no longer preserves the fine-grained structure required by faithful compression.

%% file: 6.5_discussion.tex
\section{Discussion: Hallucination or Fidelity Failure?}
\label{sec:discussion_hallucination}

\definecolor{caseyellow}{RGB}{255,211,48}
\newcommand{\casehl}[1]{\begingroup\setlength{\fboxsep}{0.8pt}\colorbox{caseyellow}{#1}\endgroup}

\begin{table}[!ht]
\vspace{-0.08in}
\footnotesize
\centering
\setlength{\tabcolsep}{3pt}
\renewcommand{\arraystretch}{1.04}
\vskip -0.06in
\caption{\textbf{Fidelity failures vs. hallucination.}}
\label{tab:hallucination_fidelity_case}
\vspace{-0.06in}
\begin{tabular}{@{}p{0.06\textwidth}p{0.33\textwidth}p{0.28\textwidth}p{0.26\textwidth}@{}}
\toprule
\rowcolor{black!6}
\textbf{} & \makecell[c]{\textit{Hallucination}} & \makecell[c]{\textit{Knowledge overwrite}} & \makecell[c]{\textit{Semantic drift}} \\
\midrule
\textbf{Source}
& The drug reduced tumor growth in mice.\newline Query: Was it tested in humans?
& The \casehl{white strawberry} was sliced for the tart.
& The \casehl{doctor} called the \casehl{nurse} after the patient fainted. \\
\addlinespace[1pt]
\textbf{Output}
& Yes, the drug succeeded in a 2023 \casehl{human clinical trial}.
& The \casehl{red strawberry} was sliced for the tart.
& The \casehl{nurse} called the \casehl{doctor} after the patient fainted. \\
\bottomrule
\end{tabular}
\vspace{-0.1in}
\end{table}

% Although these errors can look similar at the surface, their reference standards differ. \textbf{Hallucination} is an output-level grounding error. Context compression is a \textbf{lossy communication} problem: the source $\mathbf{x}$ is transmitted through memory $\mathbf{Z}$ and reconstructed as $\hat{\mathbf{x}}$. A reconstruction fails whenever it loses source-specific facts or relations, even if the substitute is fluent or plausible. The Size-Fidelity Paradox therefore reflects weakened \textbf{source-faithful recovery}, not merely hallucination under a compression setting.

Although knowledge overwriting and semantic drift can superficially resemble hallucination, they describe a different failure regime. \textbf{Hallucination} is typically an output-level grounding error, where a model introduces content unsupported by evidence or inconsistent with external truth. Context compression instead defines a \textbf{lossy communication} problem: the source $\mathbf{x}$ must be transmitted through a compact memory $\mathbf{Z}$ and reconstructed as $\hat{\mathbf{x}}$. A reconstruction is erroneous whenever it fails to preserve source-specific facts or relations, even if the substituted content is fluent, plausible, or factually correct in the world. Thus, the Size-Fidelity Paradox is not simply that larger models hallucinate more under compression; rather, scaling changes how source information is represented and resolved, improving plausible reconstruction while weakening \textbf{source-faithful recovery}.
\vskip -0.1in

%% file: 7_conclusion.tex
\section{Conclusion}

This work has demonstrated that the prevailing intuition behind parameter scaling does not straightforwardly transfer to lossy context compression. 
Across experiments ranging from 0.6B to 90B model parameters, we have identified a \textit{Size-Fidelity Paradox}: larger compressors achieve lower training loss yet produce less faithful compressed representations. 
To pinpoint how fidelity erodes, we have introduced two diagnostic QA tasks that separately measure knowledge overwriting and semantic drift, while isolating the effects of model scale. 
By holding parameter count fixed and probing the compressed representations, we have demonstrated that scaling changes how source information is organized in memory. Larger compressors tend to use broader semantic memory structures and can leave more plausible reconstructions available, weakening source-specific recovery.
These diagnostics reveal fidelity degradations that standard reconstruction evaluations overlook, establishing a principled guideline for assessing context compression effectiveness.
More broadly, this work challenges the universality of scaling laws, highlighting domains where larger models may require fundamentally different design principles to achieve desired behaviors.

% This work demonstrates that the prevailing intuition behind parameter scaling does not straightforwardly transfer to lossy context compression.
% Across compressors from 0.6B to 90B parameters, we identify a \textit{Size-Fidelity Paradox}: larger compressors can optimize better and reconstruct more fluently, yet preserve source-specific content less faithfully. 
% Through diagnostic evaluation, we show that this degradation appears as knowledge overwriting and semantic drift, two failures largely hidden by standard reconstruction metrics. 
% These findings suggest that effective compression should be judged not only by surface reconstruction quality, but also by whether the compressed memory supports source-faithful recovery. More broadly, the paradox highlights a setting where the strengths gained from scale must be aligned with the fidelity demands of the task.

\section{Acknowledgements and Disclosures}
This work was supported by the AI Search Team at Baidu. We thank the team members for their valuable discussions and feedback throughout this research. We also thank the anonymous reviewers for their constructive comments, which helped improve the clarity and presentation of this work. The authors' affiliations with Baidu are disclosed in the author list. The authors declare no other competing interests.

%% file: 8_appendix.tex
\section{Limitations}
\label{app:limitations}

Our study focuses on fidelity-critical context compression in a memory-token compressor--decoder setting. Although we evaluate two model families, multiple compression rates, decoder configurations, training domains, loss weights, and downstream tasks, the conclusions are still tied to this class of compression architectures and training objectives. Other compression paradigms, such as retrieval-based compression, extractive compression, or architectures with explicit copy mechanisms, may exhibit different scaling behavior.

Our rank and entropy analyses are intended as diagnostic probes of the compressed memory $\mathbf{Z}$, rather than a complete causal account of the paradox. Effective rank helps characterize how source information is organized in memory, while conditional entropy measures how tightly the memory constrains reconstruction under teacher forcing. These quantities reveal consistent associations with knowledge overwriting and semantic drift, but they do not by themselves specify a complete intervention for restoring fidelity in large compressors. The rank-intervention results further suggest that fidelity depends on the learned source-aligned structure of the memory, not only on scalar properties such as rank.

Finally, our diagnostic QA tasks are designed to stress source-specific fidelity, especially counterfactual facts and fine-grained semantic relations. This makes them sensitive to failures that standard reconstruction metrics overlook, but they are not meant to replace all downstream evaluations. Different applications may tolerate different levels of paraphrase, abstraction, or factual substitution. A practical compressor should therefore be selected and trained according to the fidelity requirements of the target use case. Developing large-compressor training methods that explicitly strengthen source anchoring while preserving efficiency remains an important direction for future work.

\input{8.1_reproductibility}

\section{Complete Quantitative Results}
\label{app:full_results}
\input{tab/accu_all_v2_full}

\input{8.2_QA_prompt}

\section{Additional Decoder Compatibility with Large Decoders}
\label{app:large_decoder}

\input{tab/app_large_decoder}

A possible concern is that the fidelity degradation of very large compressors may come from a mismatch between a high-capacity compressor and a weaker decoder. To test this, we pair Qwen3-32B with stronger decoders at 16$\times$ compression, including LLaMA-3.2-8B and a matched Qwen3-32B decoder. As shown in Tab.~\ref{tab:large_decoder}, increasing decoder capacity improves reconstruction and knowledge-overwriting accuracy to some extent. However, Qwen3-32B paired with Qwen3-32B still remains clearly below the Qwen3-4B compressor with a Qwen3-4B decoder on all fidelity-oriented QA metrics. Moreover, increasing the decoder from LLaMA-3.2-8B to Qwen3-32B does not improve semantic-drift QA and slightly decreases it on both FineWeb and FaithEval. These results suggest that stronger decoders can smooth or improve reconstruction, but they do not eliminate the fidelity degradation of very large compressors.

\section{Per-Dimension Results of Semantic Drift QA}
\label{app:7_dims}
\input{tab/app_7dims}

\input{8.3_analysis_rank}

\input{8.4_analysis_entropy}

%% file: 8.1_reproductibility.tex
\section{Reproducibility Details}
\label{app:reproducibility}

\subsection{Training Hyperparameters}
\label{app:training_hyper}

\begin{table}[h]
\centering
\small
\setlength{\tabcolsep}{6pt}
\renewcommand{\arraystretch}{1.12}
\caption{Pretraining hyperparameters and distributed setup.}
\label{tab:reproducibility_hparams}
\begin{tabular}{@{}ll@{}}
\toprule
\textbf{Hyperparameter} & \textbf{Setting} \\
\midrule
Optimizer & AdamW \\
Learning rate & $1\times10^{-4}$ \\
Learning-rate scheduler & Cosine \\
Global batch size & 512 \\
Warmup steps & 300 \\
Max gradient norm & 0.5 \\
Epochs & 1 \\
Precision & bfloat16 \\
Trainable modules & Compressor and memory converter \\
Frozen modules & Decoder \\
Distributed setup & $2 \times 8$ A100 GPUs \\
DeepSpeed strategy & ZeRO-3 for 32B compressors; ZeRO-1 for smaller compressors \\
Evaluation interval & Every 250 optimization steps \\
Checkpoint interval & Every 750 optimization steps \\
Model selection metric & Reconstruction validation loss \\
\bottomrule
\end{tabular}
\end{table}

In distributed training, we set the global batch size to 512, computed as $2$ examples per GPU $\times$ $4$ gradient-accumulation steps $\times$ $16$ GPUs. 
For the multimodal LLaMA-3.2-11B and 90B variants, we remove the vision encoders and retain only the text-based language model backbones, ensuring a consistent text-only setting across scales.
Detailed pretraining parameters and experimental settings are summarized in Table~\ref{tab:reproducibility_hparams}.

\subsection{Memory-Token Configuration}

All experiments use 256-token compression segments. For a compression rate $r$, each segment is compressed into
\[
  m = 256 / r
\]
memory embeddings. Therefore, the $4\times$, $16\times$, and $64\times$ settings use 64, 16, and 4 memory embeddings per segment, respectively.

For each input segment, we append $m$ learnable memory query tokens to the compressor input and take the final-layer hidden states at these memory-token positions as the compressed representation. When a context contains multiple segments, each segment is compressed independently. The projected memory embeddings are then concatenated and wrapped with memory-boundary embeddings before being passed to the fixed decoder.

\subsection{Dataset Preprocessing}

We preprocess FineWeb~\cite{penedo2024fineweb} and ArxivCorpus~\cite{clement2019arxiv} with the same pipeline. Raw documents are first tokenized with the compressor tokenizer without adding extra special tokens. We then recover the text span corresponding to each selected segment and tokenize it again with the decoder tokenizer, so that the compressor input and decoder supervision are aligned to their respective vocabularies.

Each pretraining example contains two aligned text spans: the current segment and the immediately following segment. The current segment is used for reconstruction supervision, while the following segment is used for continuation supervision. This construction supports the two pretraining objectives used in the main experiments: reconstructing the compressed text and predicting future context from the compressed prefix.

\subsection{Chunking and Filtering Rules}

Documents are chunked into consecutive 256-token segments according to the compressor tokenizer. We keep only documents with at least two full segments, since continuation supervision requires both a current segment and its following segment. Formally, a document with fewer than 512 compressor tokens is discarded.

For each remaining document with token sequence length $L$, we generate training pairs from adjacent chunks:
\[
  (C_j, C_{j+1}), \quad j = 0,\ldots,\left\lfloor L/256 \right\rfloor - 2 .
\]
Here $C_j$ is the segment to be compressed and reconstructed, and $C_{j+1}$ is the continuation target. Any trailing tokens that do not form a complete 256-token segment are discarded. The same chunking and filtering rules are used across all compressor scales and compression rates, ensuring that performance differences are not caused by different data construction procedures.

%% file: tab/accu_all_v2_full.tex
\begin{table*}[!htbp]
\scriptsize
\renewcommand{\arraystretch}{1.02}
\setlength{\tabcolsep}{1.1pt}
% \definecolor{accuavgshade}{RGB}{242,242,242}
\definecolor{accuavgshade}{RGB}{240,243,248}
\definecolor{academicblue}{RGB}{0,92,175}
\definecolor{academicred}{RGB}{180,35,35}
\providecommand{\bestdelta}{\textemdash\textemdash}
\providecommand{\accuavgcell}[1]{\cellcolor{accuavgshade}{#1}}
\providecommand{\accuavgredcell}[1]{\cellcolor{accuavgshade}{\textcolor{academicred}{#1}}}
\providecommand{\accudeltacell}[1]{\cellcolor{accuavgshade}{\textcolor{academicblue}{#1}}}
\providecommand{\accudeltaredcell}[1]{\cellcolor{accuavgshade}{\textcolor{academicred}{#1}}}
\providecommand{\accubestdeltacell}{\cellcolor{accuavgshade}{\bestdelta}}

\centering
% \vskip -0.4in
\caption{\textbf{Complete Quantitative Evaluation of the Size-Fidelity Paradox across Qwen and LLaMA Families (0.6B--90B).}
We report surface reconstruction metrics and two diagnostic QA evaluations under varying compression rates.
\textbf{Bold} marks the best model within each family/compression group.
Shaded Avg./\textcolor{academicblue}{$\Delta$} columns show category averages and relative percentage drops from the group best; \textcolor{academicred}{red} marks the worst QA Avg./drop.}
\label{tab:accuracy_full}
\vskip 0.04in

\resizebox{0.99\textwidth}{!}{%
\begin{tabular}{@{}c c c c c c c c c c c c c c c c c@{}}
\toprule
\multirow{2}{*}{\raisebox{-0.7\height}{\makecell{Compression\\Rate}}}
& \multirow{2}{*}{Model}
& \multirow{2}{*}{\centering \#Params}
& \multicolumn{6}{c}{Reconstruction}
& \multicolumn{4}{c}{\textit{Knowledge Overwriting}}
& \multicolumn{4}{c}{\textit{Semantic Drift}} \\
\cmidrule(r){4-9}\cmidrule(r){10-13}\cmidrule(l){14-17}
& & &
\makecell{BLEU}
& \makecell{ROUGE-L}
& \makecell{chrF++}
& \makecell{BERTScore}
& \accuavgcell{Avg.}
& \accudeltacell{$\Delta$}
& FaithEval & ConflictQA & \accuavgcell{Avg.} & \accudeltacell{$\Delta$}
& FineWeb & FaithEval & \accuavgcell{Avg.} & \accudeltacell{$\Delta$} \\
\midrule

\multirow{9}{*}{4$\times$} & \multirow{5}{*}{Qwen 3} & 0.6b
& 0.95 & 0.97 & 0.98 & 0.997 & \accuavgcell{0.97} & \accudeltacell{(-1.02\%)}
& 0.71 & \textbf{0.95} & \accuavgcell{0.83} & \accudeltacell{(-1.19\%)}
& 0.83 & \textbf{0.83} & \accuavgcell{\textbf{0.83}} & \accubestdeltacell \\
& & 1.7b
& 0.94 & 0.96 & 0.98 & 0.995 & \accuavgcell{0.97} & \accudeltacell{(-2.04\%)}
& \textbf{0.76} & 0.89 & \accuavgcell{0.83} & \accudeltacell{(-1.19\%)}
& 0.83 & 0.81 & \accuavgcell{0.82} & \accudeltacell{(-1.20\%)} \\
& & 4b
& 0.94 & 0.96 & 0.98 & 0.996 & \accuavgcell{0.97} & \accudeltacell{(-2.04\%)}
& 0.72 & \textbf{0.95} & \accuavgcell{\textbf{0.84}} & \accubestdeltacell
& \textbf{0.87} & 0.79 & \accuavgcell{\textbf{0.83}} & \accubestdeltacell \\
& & 8b
& 0.94 & 0.96 & 0.97 & 0.995 & \accuavgcell{0.97} & \accudeltacell{(-2.04\%)}
& 0.74 & 0.91 & \accuavgcell{0.83} & \accudeltacell{(-1.19\%)}
& 0.79 & \textbf{0.83} & \accuavgcell{0.81} & \accudeltacell{(-2.41\%)} \\
& & 32b
& \textbf{0.97} & \textbf{0.98} & \textbf{0.99} & \textbf{0.998} & \accuavgcell{\textbf{0.98}} & \accubestdeltacell
& 0.68 & 0.80 & \accuavgredcell{0.74} & \accudeltaredcell{(-11.90\%)}
& 0.60 & 0.78 & \accuavgredcell{0.69} & \accudeltaredcell{(-16.87\%)} \\
\cmidrule{2-17}
& \multirow{4}{*}{LLaMA 3.2} & 1b
& 0.94 & 0.96 & 0.97 & 0.995 & \accuavgcell{0.97} & \accudeltacell{(-2.02\%)}
& \textbf{0.77} & 0.88 & \accuavgcell{\textbf{0.83}} & \accubestdeltacell
& 0.78 & \textbf{0.85} & \accuavgcell{\textbf{0.82}} & \accubestdeltacell \\
& & 3b
& 0.90 & 0.93 & 0.95 & 0.992 & \accuavgcell{0.94} & \accudeltacell{(-4.04\%)}
& 0.74 & \textbf{0.91} & \accuavgcell{\textbf{0.83}} & \accubestdeltacell
& \textbf{0.79} & 0.84 & \accuavgcell{\textbf{0.82}} & \accubestdeltacell \\
& & 11b
& 0.92 & 0.94 & 0.97 & 0.993 & \accuavgcell{0.96} & \accudeltacell{(-3.03\%)}
& 0.65 & 0.80 & \accuavgcell{0.73} & \accudeltacell{(-12.05\%)}
& 0.64 & 0.76 & \accuavgcell{0.70} & \accudeltacell{(-14.63\%)} \\
& & 90b
& \textbf{0.97} & \textbf{0.99} & \textbf{0.99} & \textbf{0.999} & \accuavgcell{\textbf{0.99}} & \accubestdeltacell
& 0.60 & 0.81 & \accuavgredcell{0.71} & \accudeltaredcell{(-14.46\%)}
& 0.59 & 0.68 & \accuavgredcell{0.64} & \accudeltaredcell{(-21.95\%)} \\

\cmidrule{1-17}

\multirow{9}{*}{16$\times$} & \multirow{5}{*}{Qwen 3} & 0.6b
& 0.83 & 0.86 & 0.89 & 0.986 & \accuavgcell{0.89} & \accudeltacell{(-3.26\%)}
& 0.70 & \textbf{0.94} & \accuavgcell{0.82} & \accudeltacell{(-1.20\%)}
& 0.75 & 0.84 & \accuavgcell{0.80} & \accudeltacell{(-3.61\%)} \\
& & 1.7b
& \textbf{0.86} & 0.89 & 0.93 & 0.988 & \accuavgcell{0.92} & \accudeltacell{(-1.09\%)}
& \textbf{0.75} & 0.88 & \accuavgcell{0.82} & \accudeltacell{(-1.20\%)}
& 0.79 & \textbf{0.86} & \accuavgcell{\textbf{0.83}} & \accubestdeltacell \\
& & 4b
& \textbf{0.86} & \textbf{0.90} & \textbf{0.95} & 0.988 & \accuavgcell{\textbf{0.92}} & \accubestdeltacell
& 0.71 & \textbf{0.94} & \accuavgcell{\textbf{0.83}} & \accubestdeltacell
& \textbf{0.82} & 0.83 & \accuavgcell{\textbf{0.83}} & \accubestdeltacell \\
& & 8b
& 0.84 & 0.87 & 0.90 & 0.986 & \accuavgcell{0.90} & \accudeltacell{(-3.26\%)}
& 0.72 & 0.89 & \accuavgcell{0.81} & \accudeltacell{(-2.41\%)}
& 0.73 & 0.79 & \accuavgcell{0.76} & \accudeltacell{(-8.43\%)} \\
& & 32b
& 0.85 & 0.87 & 0.92 & \textbf{0.991} & \accuavgcell{0.91} & \accudeltacell{(-2.17\%)}
& 0.61 & 0.72 & \accuavgredcell{0.67} & \accudeltaredcell{(-19.28\%)}
& 0.57 & 0.74 & \accuavgredcell{0.66} & \accudeltaredcell{(-20.48\%)} \\
\cmidrule{2-17}
& \multirow{4}{*}{LLaMA 3.2} & 1b
& 0.82 & 0.85 & 0.89 & 0.983 & \accuavgcell{0.89} & \accudeltacell{(-5.38\%)}
& 0.72 & 0.84 & \accuavgcell{0.78} & \accudeltacell{(-4.88\%)}
& \textbf{0.75} & \textbf{0.90} & \accuavgcell{\textbf{0.83}} & \accubestdeltacell \\
& & 3b
& 0.81 & 0.85 & 0.88 & 0.982 & \accuavgcell{0.88} & \accudeltacell{(-5.38\%)}
& \textbf{0.73} & \textbf{0.90} & \accuavgcell{\textbf{0.82}} & \accubestdeltacell
& \textbf{0.75} & \textbf{0.90} & \accuavgcell{\textbf{0.83}} & \accubestdeltacell \\
& & 11b
& 0.85 & 0.89 & 0.92 & 0.986 & \accuavgcell{0.91} & \accudeltacell{(-2.15\%)}
& 0.63 & 0.77 & \accuavgcell{0.70} & \accudeltacell{(-14.63\%)}
& 0.61 & 0.73 & \accuavgcell{0.67} & \accudeltacell{(-19.28\%)} \\
& & 90b
& \textbf{0.86} & \textbf{0.91} & \textbf{0.96} & \textbf{0.993} & \accuavgcell{\textbf{0.93}} & \accubestdeltacell
& 0.55 & 0.74 & \accuavgredcell{0.65} & \accudeltaredcell{(-20.73\%)}
& 0.56 & 0.63 & \accuavgredcell{0.60} & \accudeltaredcell{(-27.71\%)} \\

\cmidrule{1-17}

\multirow{9}{*}{64$\times$} & \multirow{5}{*}{Qwen 3} & 0.6b
& 0.17 & 0.31 & 0.44 & 0.872 & \accuavgcell{0.45} & \accudeltacell{(-22.41\%)}
& 0.44 & 0.59 & \accuavgredcell{0.52} & \accudeltaredcell{(-17.74\%)}
& 0.31 & 0.39 & \accuavgcell{0.35} & \accudeltacell{(-20.45\%)} \\
& & 1.7b
& 0.21 & 0.39 & 0.49 & 0.880 & \accuavgcell{0.49} & \accudeltacell{(-13.79\%)}
& 0.51 & 0.60 & \accuavgcell{0.56} & \accudeltacell{(-11.29\%)}
& 0.35 & 0.42 & \accuavgcell{0.39} & \accudeltacell{(-11.36\%)} \\
& & 4b
& 0.27 & 0.44 & 0.58 & 0.894 & \accuavgcell{0.55} & \accudeltacell{(-5.17\%)}
& \textbf{0.54} & \textbf{0.70} & \accuavgcell{\textbf{0.62}} & \accubestdeltacell
& \textbf{0.41} & 0.46 & \accuavgcell{\textbf{0.44}} & \accubestdeltacell \\
& & 8b
& 0.28 & 0.46 & \textbf{0.59} & 0.925 & \accuavgcell{0.56} & \accudeltacell{(-1.72\%)}
& \textbf{0.54} & 0.67 & \accuavgcell{0.61} & \accudeltacell{(-3.23\%)}
& 0.38 & \textbf{0.49} & \accuavgcell{\textbf{0.44}} & \accubestdeltacell \\
& & 32b
& \textbf{0.29} & \textbf{0.49} & \textbf{0.59} & \textbf{0.930} & \accuavgcell{\textbf{0.58}} & \accubestdeltacell
& 0.47 & 0.56 & \accuavgredcell{0.52} & \accudeltaredcell{(-17.74\%)}
& 0.27 & 0.35 & \accuavgredcell{0.31} & \accudeltaredcell{(-29.55\%)} \\
\cmidrule{2-17}
& \multirow{4}{*}{LLaMA 3.2} & 1b
& 0.19 & 0.30 & 0.38 & 0.878 & \accuavgcell{0.44} & \accudeltacell{(-30.65\%)}
& 0.49 & 0.57 & \accuavgredcell{0.53} & \accudeltaredcell{(-14.52\%)}
& 0.33 & 0.44 & \accuavgcell{0.39} & \accudeltacell{(-7.32\%)} \\
& & 3b
& 0.29 & 0.50 & 0.62 & 0.919 & \accuavgcell{0.58} & \accudeltacell{(-6.45\%)}
& \textbf{0.55} & \textbf{0.68} & \accuavgcell{\textbf{0.62}} & \accubestdeltacell
& \textbf{0.35} & \textbf{0.47} & \accuavgcell{\textbf{0.41}} & \accubestdeltacell \\
& & 11b
& 0.30 & 0.53 & 0.63 & 0.927 & \accuavgcell{0.60} & \accudeltacell{(-4.84\%)}
& 0.51 & 0.63 & \accuavgcell{0.57} & \accudeltacell{(-8.06\%)}
& 0.24 & 0.33 & \accuavgcell{0.29} & \accudeltacell{(-31.71\%)} \\
& & 90b
& \textbf{0.33} & \textbf{0.55} & \textbf{0.67} & \textbf{0.944} & \accuavgcell{\textbf{0.62}} & \accubestdeltacell
& 0.45 & 0.61 & \accuavgredcell{0.53} & \accudeltaredcell{(-14.52\%)}
& 0.24 & 0.31 & \accuavgredcell{0.28} & \accudeltaredcell{(-34.15\%)} \\

\bottomrule
\end{tabular}%
}

% \vskip -0.2in
\end{table*}

%% file: 8.2_QA_prompt.tex
\section{Diagnostic QA Generation Prompts and Filtering Criteria}
\label{app:prompts}

% To systematically dissect the fidelity limitations of context compression, we construct two diagnostic QA datasets targeting the two failure modes studied in the main paper: \textit{Knowledge Overwriting} and \textit{Semantic Drift}. We employed a robust offline language model~(DeepSeek-R1) to generate high-quality, targeted QA pairs from the source corpora (including FaithEval and FineWeb). 
This section details the prompt engineering strategies used to elicit diagnostic samples, as well as the filtering procedures used to obtain the final datasets. The prompts are designed to strictly enforce the generation of questions that probe the model's reliance on parametric priors versus faithful context reconstruction. Task-specific filters further remove overly easy samples, focusing the evaluation on cases where source fidelity is more likely to be challenged.

\subsection{Semantic Drift QA Generation Prompt}
\label{app:prompt_semantic}
To quantify \textit{Semantic Drift}, we require questions that demand precise structural preservation rather than mere topical understanding. The prompt for this task directs the offline model to generate questions focusing on fine-grained semantic dimensions—such as role binding, modifier scope, and predicate exactness. Crucially, the prompt imposes a constraint that the ground-truth answer must be an exact substring of the original text, ensuring that any subtle paraphrasing or relational distortion in the reconstructed context is penalized.

\begin{researchquestion}
\textbf{Semantic Drift QA Generation} \\
You are an expert dataset writer building a QA suite to detect TOPIC/ENTITY DRIFT after compression+reconstruction.\\
\\
\textbf{GOAL} \\
Generate questions that will fail if the reconstructed text:\\
- shifts the main topic, scope or subject,\\
- replaces specific entities with generic categories (over-generalization),\\
- changes the stated actions or relationships, such as rewriting key verbs/relations in a way that changes meaning (relationship drift),\\
- drops list items or swaps roles/relationships (who did what to whom),\\
- breaks cross-sentence relations (cause/effect, contrast, concession, example, and progression are weakened, reversed, or rewritten).\\
\\
\textbf{IMPORTANT:} We are NOT adding distractor sentences. We want high-coverage, high-precision QA over the given CONTEXT.\\
\\
\textbf{HARD RULES} \\
- Output EXACTLY 20 QA items as JSON only (no extra text).\\
- 15 items must be answerable from CONTEXT.\\
\hspace*{1em} - For answerable items, the answer MUST be an exact substring from CONTEXT (verbatim, case/punctuation preserved).\\
\hspace*{1em} - Provide evidence\_span: a short quote from CONTEXT that contains the answer verbatim.\\
\hspace*{1em} - Answers must be SPECIFIC: do NOT use pronouns-only answers (it/they/this/that/he/she). Prefer named entities or concrete noun phrases.\\
- 5 items must be unanswerable but plausible and closely related (to detect drift-to-generic or invented details).\\
\hspace*{1em} - For these, set answer="UNANSWERABLE" and evidence\_span="".\\
\\
\textbf{COVERAGE \& DIVERSITY REQUIREMENTS (for the 15 answerable)} \\
- Make sure answers cover the entire context:\\
\hspace*{1em} - $\geq$4 answers from the first third of CONTEXT,\\
\hspace*{1em} - $\geq$4 from the middle third,\\
\hspace*{1em} - $\geq$4 from the last third.\\
- Include the following question types (minimum counts):\\
\hspace*{1em} 1) main\_topic ($\geq$1): asks for the central subject/entity or the discussed domain/category.\\
\hspace*{1em} 2) entity\_list ($\geq$2): asks for lists or multiple items mentioned (answer is the exact list span).\\
\hspace*{1em} 3) predicate\_exact ($\geq$1): asks for the EXACT verb/relational phrase used.\\
\hspace*{2em} - The gold answer must include the predicate words, not just an entity.\\
\hspace*{1em} 4) relation\_anchor ($\geq$1): asks about explicit relations: reason, contrast, cause/effect, “instead/however”, “to do X, they do Y”.\\
\hspace*{2em} - The answer should be the specific clause/phrase that states the relation (short span).\\
\hspace*{1em} 5) coreference\_link ($\geq$2): asks what a pronoun/this/that/these refers to, answered by the antecedent noun phrase in CONTEXT.\\
\hspace*{1em} 6) role\_binding ($\geq$2): asks who performs an action or what is affected (forces correct role assignment).\\
\hspace*{1em} 7) modifier\_scope ($\geq$2): asks for extracting key modifiers that often get generalized away (category qualifiers, degree/adverb, constraints).\\
\\
\textbf{QUALITY CONSTRAINTS} \\
- Keep each question unambiguous with a single correct answer span.\\
- Avoid trivial questions that can be answered by many spans (e.g., “What is mentioned?”).\\
- Avoid yes/no questions unless the answer span is a clear “yes/no” phrase in CONTEXT (rare).\\
\\
\textbf{OUTPUT JSON FORMAT} \\
\{\\
  \hspace*{1em} "qas": [\\
    \hspace*{2em} \{\\
      \hspace*{3em} "id": "T01",\\
      \hspace*{3em} "type": "answerable" | "unanswerable",\\
      \hspace*{3em} "probe": "main\_topic" | "entity\_list" | "predicate\_exact" | "relation\_anchor" | "coreference\_link" | "role\_binding" | "modifier\_scope",\\
      \hspace*{3em} "question": "...",\\
      \hspace*{3em} "answer": "...",\\
      \hspace*{3em} "evidence\_span": "..."\\
    \hspace*{2em} \}\\
  \hspace*{1em} ]\\
\}\\
\\
\textbf{CONTEXT} \\
\{CONTEXT\}\\
\\
\textbf{SELF-CHECK (must satisfy before output)} \\
- For every answerable item: answer string appears exactly in CONTEXT.\\
- For every unanswerable item: CONTEXT does not explicitly contain the answer.\\
- Ensure all probe-type minimum counts are met.
\end{researchquestion}

\subsection{Knowledge Overwriting QA Generation Prompt~(optional)}
\label{app:prompt_knowledge}
The primary objective of the \textit{Knowledge Overwriting} task is to evaluate whether the compressor retains source-specific counterfactuals or succumbs to its own internal world knowledge. Consequently, the prompt instructs the model to identify conflict-inducing facts within the text (e.g., a counterfactual birth location or an inverted causal relationship) and generate questions where the correct answer must be derived exclusively from the provided context, directly contradicting common priors.

\begin{researchquestion}
\textbf{Knowledge Overwriting QA Generation} \\
You are a high-precision question writer for measuring whether a reconstructed text stays faithful to the given context even when the context contradicts common knowledge.\\
\\
\textbf{TASK} \\
Given CONTEXT, generate QA items that:\\
- Anchor on statements that may conflict with typical world knowledge, stereotypes, or common textbook facts.\\
- The gold answers must follow CONTEXT (contextual truth), not real-world truth.\\
\\
\textbf{RULES} \\
- Write questions in the same language as CONTEXT.\\
- Produce exactly 12 QA items in JSON (no extra text).\\
- 9 items must be answerable from CONTEXT with EXACT-SUBSTRING answers (verbatim).\\
- 3 items must be unanswerable (plausible "world-knowledge" questions that readers might assume, but CONTEXT does not state).\\
- Among the 9 answerable items:\\
  \hspace*{1em} - At least 5 must be marked "world\_conflict": true\\
    \hspace*{2em} * world\_conflict means: a typical educated reader might expect a different answer from general knowledge, \\
    \hspace*{2em} BUT the answer is explicitly given in CONTEXT.\\
  \hspace*{1em} - At least 3 must include named entities / numbers / dates if present.\\
  \hspace*{1em} - At least 3 must test causal or procedural relations (X leads to Y) to detect "corrective rewriting".\\
- For each answerable item, provide:\\
  \hspace*{1em} - answer (verbatim substring)\\
  \hspace*{1em} - evidence\_span (short quote containing answer)\\
  \hspace*{1em} - world\_conflict (true/false)\\
  \hspace*{1em} - conflict\_note: 1 short sentence explaining what common belief it might contradict (do NOT correct it).\\
\\
\textbf{OUTPUT FORMAT (JSON)} \\
\{\\
  \hspace*{1em} "qas": [\\
    \hspace*{2em} \{\\
      \hspace*{3em} "id": "K1",\\
      \hspace*{3em} "type": "answerable" | "unanswerable",\\
      \hspace*{3em} "world\_conflict": true | false,\\
      \hspace*{3em} "conflict\_note": "...",\\
      \hspace*{3em} "question": "...",\\
      \hspace*{3em} "answer": "...",\\
      \hspace*{3em} "evidence\_span": "..."\\
    \hspace*{2em} \}\\
  \hspace*{1em} ]\\
\}\\
\\
\textbf{CONTEXT} \\
\{CONTEXT\}\\
\\
\textbf{SELF-CHECK} \\
- Do not “fix” the context. Gold answers must reflect CONTEXT exactly.\\
- For answerable: answer must be an exact substring in CONTEXT.\\
- For unanswerable: ensure CONTEXT does not contain the answer.
\end{researchquestion}

\subsection{Filtering Criteria for Diagnostic QA}
\label{app:qa_filtering}

We apply task-specific filtering to remove overly easy samples and focus the evaluation on cases that better expose fidelity failures. Filtering is performed with Qwen3-0.6B, 1.7B, 4B, 8B, and 32B compressors at 16$\times$ compression rate. Each source context is compressed and reconstructed with the fixed decoder, and Exact Match~(EM) is computed against the source-consistent gold answer after standard normalization.

\textbf{Knowledge Overwriting QA.}
Each context corresponds to one diagnostic triple $(q, a^{\rm gold}, \mathcal{A}^{\rm alt})$, where $a^{\rm gold}$ is the source-consistent answer and $\mathcal{A}^{\rm alt}$ contains plausible prior-based alternatives. We discard a candidate if all five Qwen3 compressors achieve EM $>0.8$. This removes cases that are faithfully recovered across scales and retains samples more likely to reveal knowledge overwriting.

\textbf{Semantic Drift QA.}
Each context is paired with 20 QA pairs covering the semantic dimensions in Sec.~\ref{subsec:drift}. For each compressor, we average EM over the 20 QA pairs for that context, and discard the context if all five compressors achieve average EM $>0.8$. This filters out structurally easy contexts and keeps examples that better test fine-grained relational preservation.

%% file: tab/app_large_decoder.tex
\begin{table}[!ht]
\centering
\footnotesize
\renewcommand{\arraystretch}{1.05}
\setlength{\tabcolsep}{3.2pt}
\caption{Additional decoder compatibility with large decoders. We evaluate Qwen3 compressors at 16$\times$ compression. ``Recons.'' denotes reconstruction accuracy on FineWeb. Numbers in parentheses indicate changes relative to the decoder configuration specified in the row label.}
\label{tab:large_decoder}
\vskip 0.04in
\resizebox{\linewidth}{!}{%
\begin{tabular}{@{}llccccc@{}}
\toprule
\multirow{2}{*}{Compressor} 
& \multirow{2}{*}{Decoder}
& \multicolumn{1}{c}{Reconstruction}
& \multicolumn{2}{c}{QA(i): \textit{Knowledge Overwriting}}
& \multicolumn{2}{c}{QA(ii): \textit{Semantic Drift}} \\
\cmidrule(lr){3-3}\cmidrule(lr){4-5}\cmidrule(lr){6-7}
& & FineWeb
& FaithEval & ConflictQA
& FineWeb & FaithEval \\
\midrule
Qwen3-4B  & Qwen3-4B     & \textbf{0.79} & \textbf{0.66} & \textbf{0.81} & \textbf{0.63} & \textbf{0.77} \\
Qwen3-32B & Qwen3-4B~(\textit{reference})     & 0.80 & 0.59 & 0.69 & 0.52 & 0.68 \\
Qwen3-32B & LLaMA-3.2-8B~\blue{(vs.~4B)} 
& 0.85 {\blue{\scriptsize(+0.05)}} 
& 0.61 {\blue{\scriptsize(+0.02)}} 
& 0.72 {\blue{\scriptsize(+0.03)}} 
& 0.57 {\blue{\scriptsize(+0.05)}} 
& 0.74 {\blue{\scriptsize(+0.06)}} \\

Qwen3-32B & Qwen3-32B~\blue{(vs.~8B)}    
& \textbf{0.86} {\blue{\scriptsize(+0.01)}} 
& 0.63 {\blue{\scriptsize(+0.02)}} 
& 0.74 {\blue{\scriptsize(+0.02)}} 
& 0.55 {\gray{\scriptsize(-0.02)}} 
& 0.70 {\gray{\scriptsize(-0.04)}} \\
\bottomrule
\end{tabular}
}
\vskip -0.04in
\end{table}

%% file: tab/app_7dims.tex
\definecolor{sectionbg}{RGB}{228, 229, 252}

\begin{table*}[!ht]
\small
\renewcommand{\arraystretch}{1.02}
\setlength{\tabcolsep}{6pt}
\centering

\vskip 0.05in
\caption{\textbf{Per-dimension semantic-drift QA results on FineWeb and FaithEval.}
Each block corresponds to one dataset. The row \#QA pairs reports the number of generated QA pairs for each semantic-drift dimension and the total number across all dimensions; the remaining rows report QA accuracy (\%) for each dimension and the corresponding overall micro-average.}
\label{tab:semantic_drift_combined}
\vskip 0.04in

\resizebox{0.99\textwidth}{!}{%
\begin{tabular}{@{}c c c @{\hspace{3pt}} c c c c c c c c@{}}
\toprule
\multirow{2}{*}{\raisebox{-0.7\height}{\makecell{Compress.\\Rate}}}
& \multirow{2}{*}{Model}
& \multirow{2}{*}{\centering \#Params}
& \multicolumn{8}{c}{Per-dimension \textbf{\textit{Semantic-Drift}} QA results} \\
\cmidrule(r){4-11}
& & & {\centering Main topic}
& {\centering Entity list} 
& {\centering Predicate exact}
& {\centering Relation anchor} 
& {\centering Coreference link}
& {\centering Role binding}
& {\centering Modifier scope}
& {\centering \textbf{Average}}\\
\midrule

\rowcolor{sectionbg}
\multicolumn{11}{c}{\textbf{\textit{(i) Dataset: FineWeb}}} \\
\midrule

\multicolumn{3}{c}{\#QA pairs} & 
1029 & 1447 & 1091 & 1104 & 1741 & 1180 & 1388 & 8980(all)\\
\midrule

\multirow{9}{*}{4$\times$}    
& \multirow{5}{*}{Qwen 3}        
& 0.6b  & 0.7839 & \textbf{0.9031} & 0.9105 & 0.7399 & 0.7624 & 0.8412 & 0.8787 & 0.8311 \\
& & 1.7b  & \textbf{0.7968} & 0.9021 & 0.8587 & 0.7734 & 0.7726 & 0.8496 & 0.8838 & 0.8341 \\
& & 4b  & 0.8649 & 0.8970 & \textbf{0.9294} & \textbf{0.8074} & \textbf{0.8282} & \textbf{0.8729} & \textbf{0.9039} & \textbf{0.8708} \\
& & 8b  & 0.7594 & 0.7794 & 0.8053 & 0.7284 & 0.7445 & 0.8651 & 0.8630 & 0.7914 \\
& & 32b  & 0.6050 & 0.6603 & 0.6165 & 0.5353 & 0.5444 & 0.6569 & 0.6114 & 0.6028 \\
\cmidrule{2-11}

& \multirow{4}{*}{LLaMA 3.2}        
& 1b  & \textbf{0.7925} & 0.7333 & 0.8375 & 0.6954 & 0.7383 & 0.8137 & \textbf{0.8561} & 0.7786 \\
& & 3b & 0.7500 & \textbf{0.8299} & \textbf{0.8537} & \textbf{0.7168} & \textbf{0.7463} & \textbf{0.8667} & 0.7784 & \textbf{0.7904} \\
& & 11b & 0.6150 & 0.6734 & 0.6331 & 0.5982 & 0.6019 & 0.6805 & 0.7010 & 0.6439 \\
& & 90b & 0.5713 & 0.6355 & 0.5884 & 0.5283 & 0.5410 & 0.6224 & 0.6381 & 0.5896 \\

\cmidrule{1-11}

\multirow{9}{*}{16$\times$}    
& \multirow{5}{*}{Qwen 3}        
& 0.6b  & 0.7140 & 0.8283 & 0.7721 & 0.6757 & 0.7135 & 0.7618 & 0.7926 & 0.7531 \\
& & 1.7b  & 0.8095 & \textbf{0.8674} & \textbf{0.8665} & 0.7327 & 0.7523 & 0.7964 & 0.7344 & 0.7919 \\
& & 4b  & \textbf{0.8226} & 0.8409 & 0.7943 & \textbf{0.7432} & \textbf{0.7839} & \textbf{0.8653} & \textbf{0.9012} & \textbf{0.8226} \\
& & 8b  & 0.7397 & 0.7984 & 0.7725 & 0.6750 & 0.6649 & 0.7274 & 0.7379 & 0.7288 \\
& & 32b  & 0.5954 & 0.5709 & 0.5953 & 0.5142 & 0.5429 & 0.6136 & 0.5917 & 0.5731 \\
\cmidrule{2-11}

& \multirow{4}{*}{LLaMA 3.2}        
& 1b  & 0.7163 & 0.8058 & 0.7312 & \textbf{0.6857} & \textbf{0.7082} & 0.7563 & \textbf{0.8143} & 0.7476 \\
& & 3b & \textbf{0.7495} & \textbf{0.8211} & \textbf{0.7723} & 0.6710 & 0.7035 & \textbf{0.7684} & 0.7565 & \textbf{0.7488} \\
& & 11b & 0.5710 & 0.6658 & 0.6465 & 0.5497 & 0.5595 & 0.5989 & 0.6611 & 0.6082 \\
& & 90b & 0.5701 & 0.5963 & 0.5783 & 0.5211 & 0.5168 & 0.5824 & 0.6162 & 0.5677 \\

\cmidrule{1-11}

\multirow{9}{*}{64$\times$}    
& \multirow{5}{*}{Qwen 3}        
& 0.6b  & 0.2998 & 0.3436 & 0.3423 & 0.2823 & 0.2988 & 0.3441 & 0.2908 & 0.3141 \\
& & 1.7b  & 0.3305 & 0.3543 & 0.3749 & 0.3170 & 0.3288 & 0.3643 & 0.3804 & 0.3499 \\
& & 4b  & \textbf{0.4079} & 0.4016 & \textbf{0.4297} & \textbf{0.3856} & \textbf{0.3764} & \textbf{0.4487} & \textbf{0.4529} & \textbf{0.4130} \\
& & 8b  & 0.3768 & \textbf{0.4086} & 0.3881 & 0.3485 & 0.3524 & 0.4109 & 0.3511 & 0.3756 \\
& & 32b  & 0.2728 & 0.2929 & 0.2861 & 0.2491 & 0.2467 & 0.2849 & 0.2911 & 0.2741 \\
\cmidrule{2-11}

& \multirow{4}{*}{LLaMA 3.2}        
& 1b  & 0.3251 & 0.3165 & 0.3638 & 0.3087 & 0.3144 & \textbf{0.3610} & 0.3436 & 0.3319 \\
& & 3b & \textbf{0.3534} & \textbf{0.3559} & \textbf{0.3818} & \textbf{0.3250} & \textbf{0.3298} & 0.3421 & \textbf{0.3624} & \textbf{0.3491} \\
& & 11b & 0.2484 & 0.2556 & 0.2658 & 0.2183 & 0.2229 & 0.2334 & 0.2674 & 0.2440 \\
& & 32b & 0.2479 & 0.2503 & 0.2486 & 0.2205 & 0.2228 & 0.2458 & 0.2404 & 0.2387 \\

\midrule
\rowcolor{sectionbg}
\multicolumn{11}{c}{\textbf{\textit{(ii) Dataset: FaithEval}}} \\
\midrule

\multicolumn{3}{c}{\#QA pairs} & 
1207 & 1973 & 1363 & 1495 & 2005 & 2019 & 1898 & 11960(all)\\
\midrule

\multirow{9}{*}{4$\times$}    
& \multirow{5}{*}{Qwen 3}        
& 0.6b  & 0.7855 & \textbf{0.9046} & \textbf{0.9120} & 0.7414 & 0.7639 & 0.8427 & \textbf{0.8656} & \textbf{0.8328} \\
& & 1.7b  & 0.7902 & 0.8213 & 0.8786 & 0.7457 & 0.7363 & \textbf{0.8826} & 0.8248 & 0.8119 \\
& & 4b  & 0.7929 & 0.8221 & 0.7992 & 0.7265 & 0.7437 & 0.7558 & 0.8558 & 0.7856 \\
& & 8b  & \textbf{0.8385} & 0.8373 & 0.8578 & \textbf{0.7744} & \textbf{0.7727} & 0.8492 & 0.8570 & 0.8262 \\
& & 32b  & 0.7773 & 0.8351 & 0.8556 & 0.6962 & 0.7306 & 0.7458 & 0.8285 & 0.7806 \\
\cmidrule{2-11}

& \multirow{4}{*}{LLaMA 3.2}        
& 1b  & 0.8452 & \textbf{0.9084} & \textbf{0.8809} & \textbf{0.7955} & 0.7757 & \textbf{0.8633} & 0.9051 & \textbf{0.8544} \\
& & 3b & \textbf{0.8509} & 0.8998 & 0.8013 & 0.7881 & \textbf{0.7962} & 0.8214 & \textbf{0.9233} & 0.8428 \\
& & 11b & 0.7463 & 0.7645 & 0.8236 & 0.7163 & 0.7199 & 0.8180 & 0.7593 & 0.7641 \\
& & 90b & 0.6475 & 0.7041 & 0.6638 & 0.6292 & 0.6331 & 0.7112 & 0.7361 & 0.6788 \\

\cmidrule{1-11}

\multirow{9}{*}{16$\times$}    
& \multirow{5}{*}{Qwen 3}        
& 0.6b  & \textbf{0.8736} & \textbf{0.9060} & 0.8498 & 0.7541 & 0.7924 & 0.9251 & 0.7792 & 0.8414 \\
& & 1.7b  & 0.8281 & 0.9048 & \textbf{0.9292} & \textbf{0.8080} & \textbf{0.8003} & \textbf{0.9362} & 0.8209 & \textbf{0.8622} \\
& & 4b  & 0.7894 & 0.8595 & 0.8647 & 0.7417 & 0.7923 & 0.8972 & \textbf{0.8690} & 0.8349 \\
& & 8b  & 0.8114 & 0.8143 & 0.8159 & 0.7228 & 0.7197 & 0.8336 & 0.7850 & 0.7855 \\
& & 32b  & 0.7359 & 0.7687 & 0.8080 & 0.6933 & 0.6782 & 0.7009 & 0.8082 & 0.7401 \\
\cmidrule{2-11}

& \multirow{4}{*}{LLaMA 3.2}        
& 1b  & \textbf{0.9285} & 0.8395 & \textbf{0.9387} & 0.8417 & 0.8462 & 0.9322 & \textbf{0.9647} & 0.8967 \\
& & 3b & 0.8502 & \textbf{0.9518} & 0.8850 & \textbf{0.8448} & \textbf{0.8471} & \textbf{0.9621} & 0.9414 & \textbf{0.9031} \\
& & 11b &0.7478 & 0.7839 & 0.7571 & 0.6882 & 0.6657 & 0.7093 & 0.7906 & 0.7339 \\
& & 90b &0.6341 & 0.6750 & 0.5954 & 0.5822 & 0.5946 & 0.6454 & 0.6799 & 0.6325 \\

\cmidrule{1-11}

\multirow{9}{*}{64$\times$}    
& \multirow{5}{*}{Qwen 3}        
& 0.6b  & 0.4038 & 0.4183 & 0.3918 & 0.3623 & 0.3587 & 0.3776 & 0.4205 & 0.3903 \\
& & 1.7b  & 0.3977 & 0.4363 & 0.4513 & 0.3888 & 0.3783 & 0.4491 & 0.4287 & 0.4194 \\
& & 4b  & 0.4659 & 0.4744 & 0.4664 & 0.4258 & 0.4318 & 0.4589 & \textbf{0.5084} & 0.4622 \\
& & 8b  & \textbf{0.4671} & \textbf{0.5323} & \textbf{0.5400} & \textbf{0.4482} & \textbf{0.4661} & \textbf{0.5342} & 0.4447 & \textbf{0.4914} \\
& & 32b  & 0.3419 & 0.3396 & 0.3693 & 0.3263 & 0.3229 & 0.3738 & 0.3697 & 0.3493 \\
\cmidrule{2-11}

& \multirow{4}{*}{LLaMA 3.2}        
& 1b  & 0.4147 & \textbf{0.4621} & 0.4604 & 0.4125 & 0.3963 & 0.4582 & 0.4688 & 0.4403 \\
& & 3b & \textbf{0.4766} & 0.4490 & \textbf{0.4634} & \textbf{0.4254} & \textbf{0.4437} & \textbf{0.5148} & \textbf{0.4988} & \textbf{0.4686} \\
& & 11b &0.3288 & 0.3388 & 0.3310 & 0.3012 & 0.2963 & 0.3558 & 0.3524 & 0.3301 \\
& & 90b &0.3043 & 0.3325 & 0.3453 & 0.2932 & 0.2929 & 0.3366 & 0.2937 & 0.3141 \\

\bottomrule
\end{tabular}%
}

\end{table*}

%% file: 8.3_analysis_rank.tex
\section{Additional Analysis of Memory Geometry}
\label{app:memory_geometry}

\subsection{Effective Rank and Hidden Dimension}
\label{app:erank_hidden}

We further examine whether the increase in effective rank is merely a consequence of larger hidden dimensions. For this analysis, memory tokens are stacked as observations and the effective rank is computed over the hidden dimension. As shown in Tab.~\ref{tab:erank_hidden}, effective rank increases with scale, but the normalized ratio $\mathrm{erank}/d$ decreases. Thus, larger compressors do not simply occupy the expanded hidden space uniformly. Instead, they form broader yet still relatively low-dimensional memory structures, consistent with the view that scaling changes how source information is organized rather than only increasing representational size.

\begin{table}[t]
\centering
\small
\caption{Effective rank normalized by hidden size.}
\label{tab:erank_hidden}
\begin{tabular}{lccc}
\toprule
Model & Hidden size $d$ & eRank & eRank/$d$ \\
\midrule
Qwen3-0.6B  & 1024 & 856.8  & 0.8455 \\
Qwen3-1.7B  & 2048 & 1032.8 & 0.5043 \\
Qwen3-4B    & 2560 & 1067.0 & 0.4168 \\
Qwen3-8B    & 4096 & 1134.4 & 0.2770 \\
Qwen3-32B   & 5120 & 1169.3 & 0.2284 \\
LLaMA-3.2-1B  & 2048 & 975.2  & 0.4762 \\
LLaMA-3.2-3B  & 3072 & 1072.6 & 0.3492 \\
LLaMA-3.2-11B & 4096 & 1149.3 & 0.2810 \\
LLaMA-3.2-90B & 8192 & 1194.4 & 0.1458 \\
\bottomrule
\end{tabular}
\end{table}

\subsection{Rank Intervention and Spectral Structure}
\label{app:rank_intervention}

\begin{figure*}[ht]
  \vskip -0.1in
  \begin{center}
    \centerline{\includegraphics[width=0.99\columnwidth]{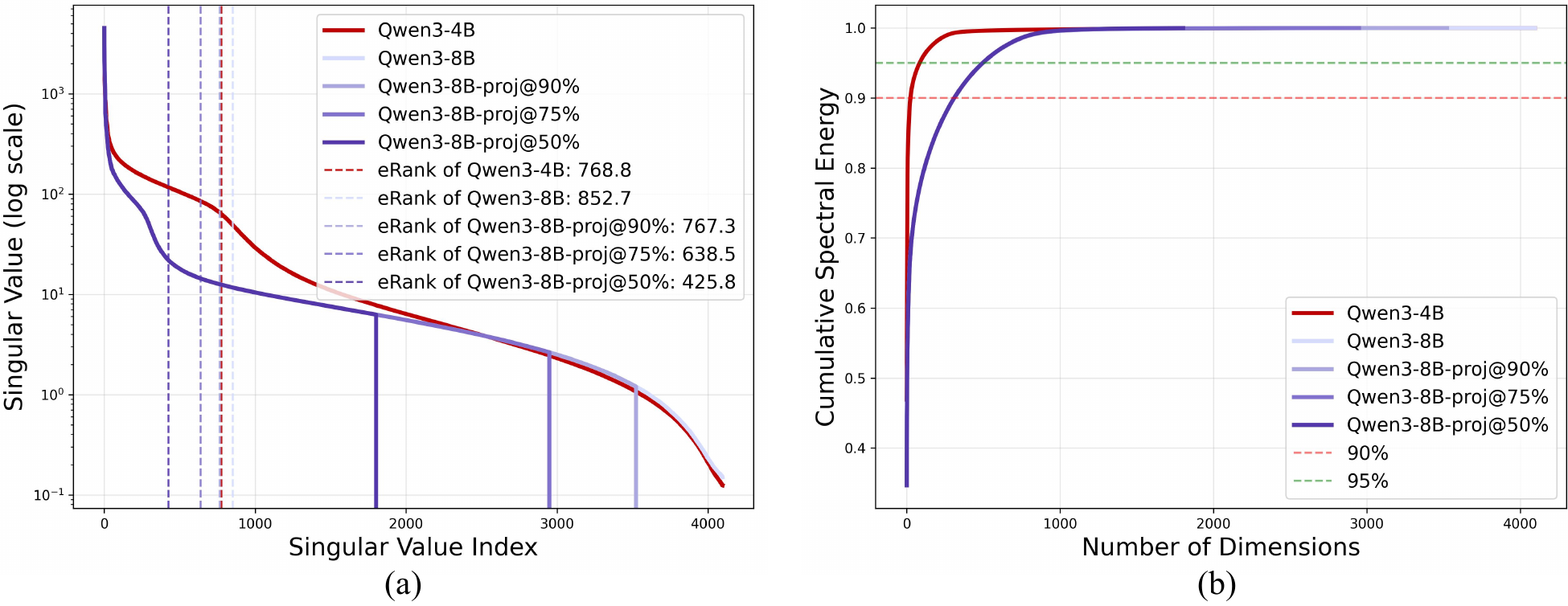}}
    \caption{\textbf{Spectral evidence for a source-sufficient regime.} Mild truncation removes low-energy tail directions, whereas stronger truncation cuts into source-bearing spectral regions.
\textbf{(a) Singular-value spectra.} The natural 8B compressor has a flatter, longer-tailed spectrum than the natural 4B model. A 90\% projection mainly trims the tail, while 75\%/50\% projections cut into the shoulder near the spectral elbow.
\textbf{(b) Cumulative spectral energy.} The 90\% projection preserves most leading spectral energy, whereas stronger projections discard increasingly larger portions of the spectrum, explaining the transition from improved fidelity to source-signal loss.}
    \label{fig:rank_intervention_spectrum}
  \end{center}
\end{figure*}

\begin{figure*}[ht]
  \vskip -0.2in
  \begin{center}
\centerline{\includegraphics[width=0.99\columnwidth]{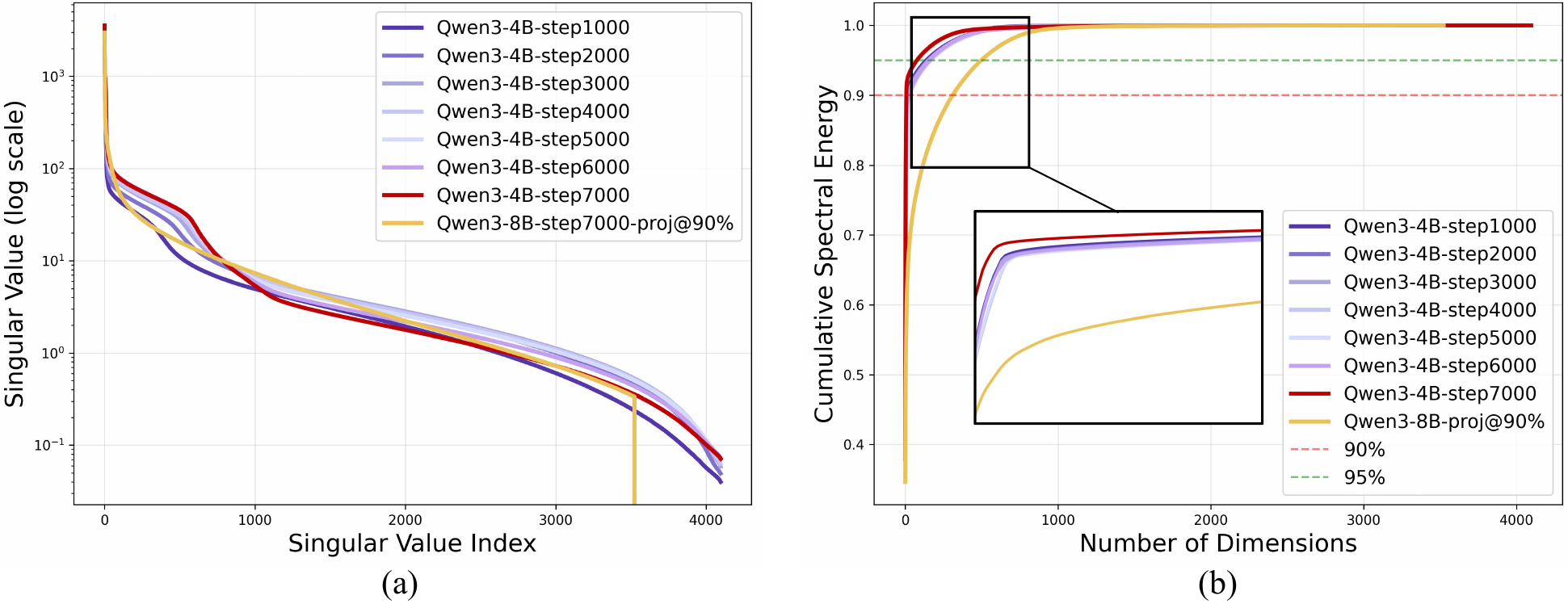}}
\caption{\textbf{Training forms the source-aligned spectral structure.}
\textbf{(a) Singular-value spectra across 4B training.}
As training proceeds, the 4B compressor evolves toward a smoother and more stable low-rank spectrum, suggesting that the source-sufficient structure is formed during optimization rather than imposed afterward. The 90\%-projected 8B approaches the late-stage 4B in spectral concentration, but remains a truncation of a different spectrum.
\textbf{(b) Cumulative spectral energy across 4B training.}
The 4B model progressively concentrates more spectral energy into a compact leading subspace, consistent with convergence toward a source-aligned memory structure. Thus, similar $eRank$ can reflect similar spectral concentration without guaranteeing the same represented subspace or reconstruction fidelity.}
    \label{fig:rank_intervention_spectrum_training}
  \end{center}
\end{figure*}

To further probe the role of memory geometry, we apply post-hoc truncated-SVD projection to Qwen3-8B memory embeddings at 16$\times$ compression. Tab.~\ref{tab:rank_intervention} and Fig.~\ref{fig:rank_intervention_spectrum}-\ref{fig:rank_intervention_spectrum_training} shows a non-monotonic pattern. Moderate truncation improves reconstruction and both fidelity-oriented QA scores, while stronger truncation sharply degrades them. This suggests that part of the spectral tail corresponds to diffuse semantic variation, but the leading subspace still contains source-bearing evidence that must be preserved.

\begin{table}[t]
\centering
\small
\caption{Rank intervention on Qwen3-8B memory embeddings at 16$\times$ compression.}
\label{tab:rank_intervention}
\begin{tabular}{lccc}
\toprule
% Model / Intervention & Reconstruction & \textit{Knowledge Overwrite} & \textit{Semmantic Drift} \\

\multirow{2}{*}{Compressor / Intervention} 
& \multicolumn{1}{c}{Reconstruction}
& \multicolumn{1}{c}{QA(i): \textit{Knowledge Overwriting}}
& \multicolumn{1}{c}{QA(ii): \textit{Semantic Drift}} \\
\cmidrule(lr){2-2}\cmidrule(lr){3-3}\cmidrule(lr){4-4}
 & FineWeb
& ConflictQA
& FineWeb \\

\midrule
Qwen3-4B & \textbf{0.8622} & \textbf{0.9414} & \textbf{0.8226} \\
Qwen3-8B \textit{(reference)} & 0.8419 & 0.8931 & 0.7288 \\
Qwen3-8B \textit{(90\%)} & \underline{0.8493} {\blue{\scriptsize(+0.0074)}} & \underline{0.9104} {\blue{\scriptsize(+0.0173)}} & \underline{0.7452} {\blue{\scriptsize(+0.0164)}} \\
Qwen3-8B \textit{(75\%)} & 0.7933 {\gray{\scriptsize(-0.0486)}} & 0.7802 {\gray{\scriptsize(-0.1129)}} & 0.6685 {\gray{\scriptsize(-0.0603)}} \\
Qwen3-8B \textit{(50\%)} & 0.7276 {\gray{\scriptsize(-0.1143)}} & 0.7313 {\gray{\scriptsize(-0.1618)}} & 0.5841 {\gray{\scriptsize(-0.1447)}} \\
\bottomrule
\end{tabular}
\end{table}

% \begin{figure*}[t]
% \centering
% \fbox{\parbox[c][1.25in][c]{0.45\linewidth}{\centering Placeholder: singular-value spectra}}
% \hfill
% \fbox{\parbox[c][1.25in][c]{0.45\linewidth}{\centering Placeholder: cumulative spectral energy}}
% \caption{Spectral view of the rank intervention. Mild truncation removes low-energy tail directions, whereas stronger truncation cuts into source-bearing spectral regions.}
% \label{fig:rank_intervention_spectrum}
% \end{figure*}

%% file: 8.4_analysis_entropy.tex
\section{Additional Analysis of Reconstruction Entropy}
\label{app:entropy_appendix}

\subsection{A Bottleneck View of Conditional Entropy}
\label{app:entropy_theory}

A principled way to interpret the entropy trend is to view context compression as source transmission under a fixed bottleneck. Let $X$ denote the source, $Z=f_\theta(X)$ the compressed memory, and $S=h(X)$ a coarse semantic abstraction of the source. Since $S$ is determined by $X$,
\[
I(X;Z)=I(S;Z)+I(X;Z\mid S).
\]
Under a fixed memory budget, increasing the amount of semantic information preserved in $Z$ does not guarantee that source-specific residual details are equally preserved. The remaining uncertainty satisfies
\[
H(X\mid Z)=H(S\mid Z)+H(X\mid S,Z),
\quad
H(X\mid S,Z)=H(X\mid S)-I(X;Z\mid S).
\]
Thus, when $Z$ captures broad semantic content but under-specifies source-specific details, multiple reconstructions can remain compatible with the same memory. The teacher-forced conditional entropy used in Sec.~\ref{subsec:entropy_analysis} provides a token-level probe of this ambiguity. Higher entropy indicates that the memory leaves more plausible continuations available, which is precisely the condition under which fine-grained semantic drift becomes more likely.

\subsection{Entropy under Loss-Weight Sweep}
\label{app:lambda_entropy}

\begin{figure*}[ht]
  \vskip 0.2in
  \begin{center}
    \centerline{\includegraphics[width=1.0\columnwidth]{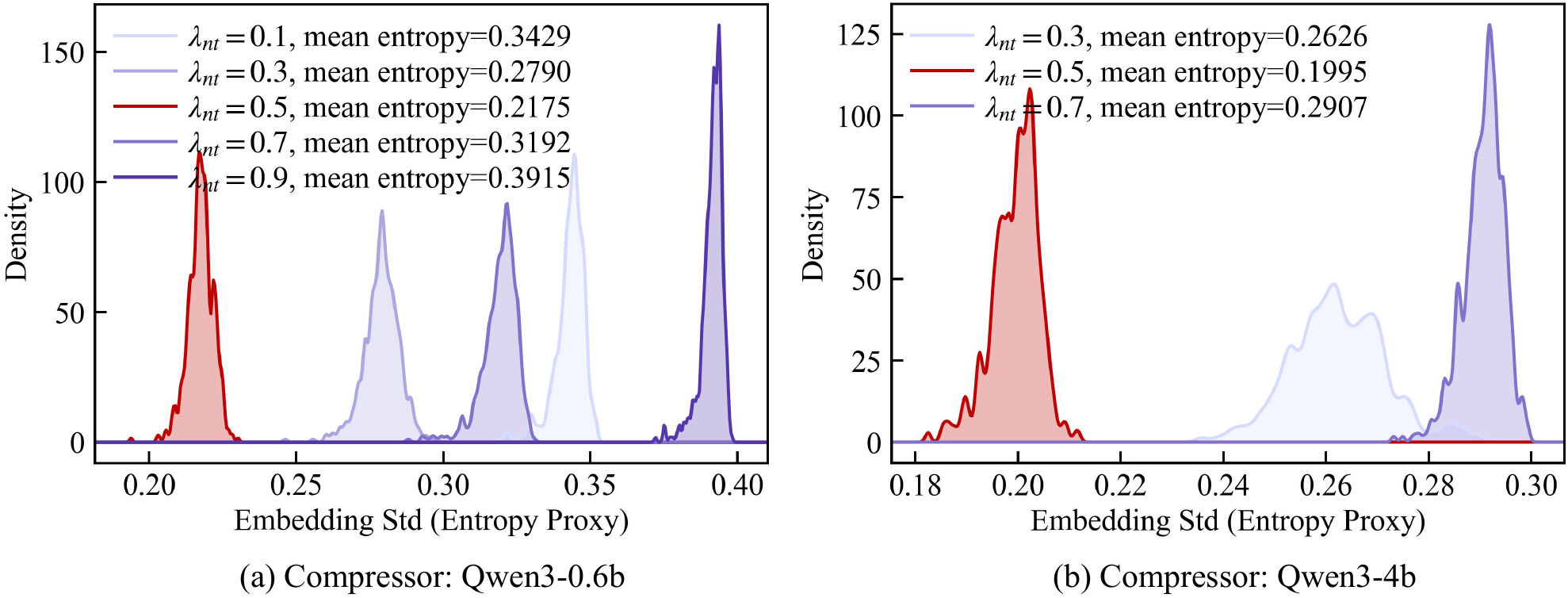}}
    \caption{\textbf{Entropy distributions under $\lambda$-sweep.} Shown are the embedding-entropy distributions of compressors trained with different $\lambda$ values ($\lambda L_{nt} + (1-\lambda)L_{re}$) for Qwen3-0.6B(a) and Qwen3-4B(b) at 4$\times$; in both cases, the mean entropy is lowest around $\lambda = 0.5$.}
    \label{fig:lambda_entropy}
  \end{center}
\end{figure*}

We also examine conditional entropy under the loss-weight sweep in Sec.~\ref{sec:loss_sweep}. Fig.~\ref{fig:lambda_entropy} shows the entropy distributions for Qwen3-0.6B and Qwen3-4B at 4$\times$ compression. In both cases, entropy is lowest near the balanced setting $\lambda=0.5$, where reconstruction and next-token prediction jointly shape the memory. Moving away from this balance broadens the conditional distribution, suggesting that reconstruction determinacy depends on how the objective allocates capacity between source recovery and general language modeling.

% \begin{figure*}[t]
% \centering
% \fbox{\parbox[c][1.25in][c]{0.45\linewidth}{\centering Placeholder: Qwen3-0.6B entropy under $\lambda$ sweep}}
% \hfill
% \fbox{\parbox[c][1.25in][c]{0.45\linewidth}{\centering Placeholder: Qwen3-4B entropy under $\lambda$ sweep}}
% \caption{Conditional entropy distributions under loss-weight sweep. The lowest entropy appears near the balanced objective, indicating more determinate reconstruction from the compressed memory.}
% \label{fig:lambda_entropy}
% \end{figure*}

%% file: checklist.tex
\newpage
\section*{NeurIPS Paper Checklist}

\begin{enumerate}

\item {\bf Claims}
    \item[] Question: Do the main claims made in the abstract and introduction accurately reflect the paper's contributions and scope?
    \item[] Answer: \answerYes{} % Replace by \answerYes{}, \answerNo{}, or \answerNA{}.
    \item[] Justification: See Sec.~\ref{intro} for the main claims and Sec.~\ref{sec:paradox} for the detailed contributions and scope.
    \item[] Guidelines:
    \begin{itemize}
        \item The answer \answerNA{} means that the abstract and introduction do not include the claims made in the paper.
        \item The abstract and/or introduction should clearly state the claims made, including the contributions made in the paper and important assumptions and limitations. A \answerNo{} or \answerNA{} answer to this question will not be perceived well by the reviewers. 
        \item The claims made should match theoretical and experimental results, and reflect how much the results can be expected to generalize to other settings. 
        \item It is fine to include aspirational goals as motivation as long as it is clear that these goals are not attained by the paper. 
    \end{itemize}

\item {\bf Limitations}
    \item[] Question: Does the paper discuss the limitations of the work performed by the authors?
    \item[] Answer: \answerYes{} % Replace by \answerYes{}, \answerNo{}, or \answerNA{}.
    \item[] Justification: See Appendix~\ref{app:limitations} for the discussion on the limitations.
    \item[] Guidelines:
    \begin{itemize}
        \item The answer \answerNA{} means that the paper has no limitation while the answer \answerNo{} means that the paper has limitations, but those are not discussed in the paper. 
        \item The authors are encouraged to create a separate ``Limitations'' section in their paper.
        \item The paper should point out any strong assumptions and how robust the results are to violations of these assumptions (e.g., independence assumptions, noiseless settings, model well-specification, asymptotic approximations only holding locally). The authors should reflect on how these assumptions might be violated in practice and what the implications would be.
        \item The authors should reflect on the scope of the claims made, e.g., if the approach was only tested on a few datasets or with a few runs. In general, empirical results often depend on implicit assumptions, which should be articulated.
        \item The authors should reflect on the factors that influence the performance of the approach. For example, a facial recognition algorithm may perform poorly when image resolution is low or images are taken in low lighting. Or a speech-to-text system might not be used reliably to provide closed captions for online lectures because it fails to handle technical jargon.
        \item The authors should discuss the computational efficiency of the proposed algorithms and how they scale with dataset size.
        \item If applicable, the authors should discuss possible limitations of their approach to address problems of privacy and fairness.
        \item While the authors might fear that complete honesty about limitations might be used by reviewers as grounds for rejection, a worse outcome might be that reviewers discover limitations that aren't acknowledged in the paper. The authors should use their best judgment and recognize that individual actions in favor of transparency play an important role in developing norms that preserve the integrity of the community. Reviewers will be specifically instructed to not penalize honesty concerning limitations.
    \end{itemize}

\item {\bf Theory assumptions and proofs}
    \item[] Question: For each theoretical result, does the paper provide the full set of assumptions and a complete (and correct) proof?
    \item[] Answer: \answerYes{} % Replace by \answerYes{}, \answerNo{}, or \answerNA{}.
    \item[] Justification:  See Sec.~\ref{sec:principle} for assumptions and proof.
    \item[] Guidelines:
    \begin{itemize}
        \item The answer \answerNA{} means that the paper does not include theoretical results. 
        \item All the theorems, formulas, and proofs in the paper should be numbered and cross-referenced.
        \item All assumptions should be clearly stated or referenced in the statement of any theorems.
        \item The proofs can either appear in the main paper or the supplemental material, but if they appear in the supplemental material, the authors are encouraged to provide a short proof sketch to provide intuition. 
        \item Inversely, any informal proof provided in the core of the paper should be complemented by formal proofs provided in appendix or supplemental material.
        \item Theorems and Lemmas that the proof relies upon should be properly referenced. 
    \end{itemize}

    \item {\bf Experimental result reproducibility}
    \item[] Question: Does the paper fully disclose all the information needed to reproduce the main experimental results of the paper to the extent that it affects the main claims and/or conclusions of the paper (regardless of whether the code and data are provided or not)?
    \item[] Answer: \answerYes{} % Replace by \answerYes{}, \answerNo{}, or \answerNA{}.
    \item[] Justification: See Appendix~\ref{app:reproducibility} for reproducibility details.
    \item[] Guidelines:
    \begin{itemize}
        \item The answer \answerNA{} means that the paper does not include experiments.
        \item If the paper includes experiments, a \answerNo{} answer to this question will not be perceived well by the reviewers: Making the paper reproducible is important, regardless of whether the code and data are provided or not.
        \item If the contribution is a dataset and\slash or model, the authors should describe the steps taken to make their results reproducible or verifiable. 
        \item Depending on the contribution, reproducibility can be accomplished in various ways. For example, if the contribution is a novel architecture, describing the architecture fully might suffice, or if the contribution is a specific model and empirical evaluation, it may be necessary to either make it possible for others to replicate the model with the same dataset, or provide access to the model. In general. releasing code and data is often one good way to accomplish this, but reproducibility can also be provided via detailed instructions for how to replicate the results, access to a hosted model (e.g., in the case of a large language model), releasing of a model checkpoint, or other means that are appropriate to the research performed.
        \item While NeurIPS does not require releasing code, the conference does require all submissions to provide some reasonable avenue for reproducibility, which may depend on the nature of the contribution. For example
        \begin{enumerate}
            \item If the contribution is primarily a new algorithm, the paper should make it clear how to reproduce that algorithm.
            \item If the contribution is primarily a new model architecture, the paper should describe the architecture clearly and fully.
            \item If the contribution is a new model (e.g., a large language model), then there should either be a way to access this model for reproducing the results or a way to reproduce the model (e.g., with an open-source dataset or instructions for how to construct the dataset).
            \item We recognize that reproducibility may be tricky in some cases, in which case authors are welcome to describe the particular way they provide for reproducibility. In the case of closed-source models, it may be that access to the model is limited in some way (e.g., to registered users), but it should be possible for other researchers to have some path to reproducing or verifying the results.
        \end{enumerate}
    \end{itemize}

\item {\bf Open access to data and code}
    \item[] Question: Does the paper provide open access to the data and code, with sufficient instructions to faithfully reproduce the main experimental results, as described in supplemental material?
    \item[] Answer: \answerYes{} % Replace by \answerYes{}, \answerNo{}, or \answerNA{}.
    \item[] Justification: We include our data in the supplementary material, covering the contexts and QA pairs for both \textit{knowledge overwriting} and \textit{semantic drift}. The code follows the PCC compressor--decoder paradigm~\cite{dai2025pretraining}, with reproducibility instructions provided in Appendix~\ref{app:reproducibility}.

    \item[] Guidelines:
    \begin{itemize}
        \item The answer \answerNA{} means that paper does not include experiments requiring code.
        \item Please see the NeurIPS code and data submission guidelines (\url{https://neurips.cc/public/guides/CodeSubmissionPolicy}) for more details.
        \item While we encourage the release of code and data, we understand that this might not be possible, so \answerNo{} is an acceptable answer. Papers cannot be rejected simply for not including code, unless this is central to the contribution (e.g., for a new open-source benchmark).
        \item The instructions should contain the exact command and environment needed to run to reproduce the results. See the NeurIPS code and data submission guidelines (\url{https://neurips.cc/public/guides/CodeSubmissionPolicy}) for more details.
        \item The authors should provide instructions on data access and preparation, including how to access the raw data, preprocessed data, intermediate data, and generated data, etc.
        \item The authors should provide scripts to reproduce all experimental results for the new proposed method and baselines. If only a subset of experiments are reproducible, they should state which ones are omitted from the script and why.
        \item At submission time, to preserve anonymity, the authors should release anonymized versions (if applicable).
        \item Providing as much information as possible in supplemental material (appended to the paper) is recommended, but including URLs to data and code is permitted.
    \end{itemize}

\item {\bf Experimental setting/details}
    \item[] Question: Does the paper specify all the training and test details (e.g., data splits, hyperparameters, how they were chosen, type of optimizer) necessary to understand the results?
    \item[] Answer: \answerYes{} % Replace by \answerYes{}, \answerNo{}, or \answerNA{}.
    \item[] Justification: We present our experiment settings and implementation details in Sec.~\ref{subsec:setup} and Appendix~\ref{app:reproducibility}.
    \item[] Guidelines:
    \begin{itemize}
        \item The answer \answerNA{} means that the paper does not include experiments.
        \item The experimental setting should be presented in the core of the paper to a level of detail that is necessary to appreciate the results and make sense of them.
        \item The full details can be provided either with the code, in appendix, or as supplemental material.
    \end{itemize}

\item {\bf Experiment statistical significance}
    \item[] Question: Does the paper report error bars suitably and correctly defined or other appropriate information about the statistical significance of the experiments?
    \item[] Answer: \answerYes{} % Replace by \answerYes{}, \answerNo{}, or \answerNA{}.
    \item[] Justification:  We provide the the statistical significance of the experiments in Sec.~\ref{sec:generalization}.
    \item[] Guidelines:
    \begin{itemize}
        \item The answer \answerNA{} means that the paper does not include experiments.
        \item The authors should answer \answerYes{} if the results are accompanied by error bars, confidence intervals, or statistical significance tests, at least for the experiments that support the main claims of the paper.
        \item The factors of variability that the error bars are capturing should be clearly stated (for example, train/test split, initialization, random drawing of some parameter, or overall run with given experimental conditions).
        \item The method for calculating the error bars should be explained (closed form formula, call to a library function, bootstrap, etc.)
        \item The assumptions made should be given (e.g., Normally distributed errors).
        \item It should be clear whether the error bar is the standard deviation or the standard error of the mean.
        \item It is OK to report 1-sigma error bars, but one should state it. The authors should preferably report a 2-sigma error bar than state that they have a 96\% CI, if the hypothesis of Normality of errors is not verified.
        \item For asymmetric distributions, the authors should be careful not to show in tables or figures symmetric error bars that would yield results that are out of range (e.g., negative error rates).
        \item If error bars are reported in tables or plots, the authors should explain in the text how they were calculated and reference the corresponding figures or tables in the text.
    \end{itemize}

\item {\bf Experiments compute resources}
    \item[] Question: For each experiment, does the paper provide sufficient information on the computer resources (type of compute workers, memory, time of execution) needed to reproduce the experiments?
    \item[] Answer: \answerYes{} % Replace by \answerYes{}, \answerNo{}, or \answerNA{}.
    \item[] Justification: We introduce the compute resources need for our work in Appendix~\ref{app:training_hyper}.
    \item[] Guidelines:
    \begin{itemize}
        \item The answer \answerNA{} means that the paper does not include experiments.
        \item The paper should indicate the type of compute workers CPU or GPU, internal cluster, or cloud provider, including relevant memory and storage.
        \item The paper should provide the amount of compute required for each of the individual experimental runs as well as estimate the total compute. 
        \item The paper should disclose whether the full research project required more compute than the experiments reported in the paper (e.g., preliminary or failed experiments that didn't make it into the paper). 
    \end{itemize}
    
\item {\bf Code of ethics}
    \item[] Question: Does the research conducted in the paper conform, in every respect, with the NeurIPS Code of Ethics \url{https://neurips.cc/public/EthicsGuidelines}?
    \item[] Answer: \answerYes{} % Replace by \answerYes{}, \answerNo{}, or \answerNA{}.
    \item[] Justification: The research conducted in the paper conform, in every respect, with the NeurIPS Code of Ethics.
    \item[] Guidelines:
    \begin{itemize}
        \item The answer \answerNA{} means that the authors have not reviewed the NeurIPS Code of Ethics.
        \item If the authors answer \answerNo, they should explain the special circumstances that require a deviation from the Code of Ethics.
        \item The authors should make sure to preserve anonymity (e.g., if there is a special consideration due to laws or regulations in their jurisdiction).
    \end{itemize}

\item {\bf Broader impacts}
    \item[] Question: Does the paper discuss both potential positive societal impacts and negative societal impacts of the work performed?
    \item[] Answer: \answerYes{} % Replace by \answerYes{}, \answerNo{}, or \answerNA{}.
    \item[] Justification: We discuss broader impacts in Sec.~\ref{sec:downstream}.
    \item[] Guidelines:
    \begin{itemize}
        \item The answer \answerNA{} means that there is no societal impact of the work performed.
        \item If the authors answer \answerNA{} or \answerNo, they should explain why their work has no societal impact or why the paper does not address societal impact.
        \item Examples of negative societal impacts include potential malicious or unintended uses (e.g., disinformation, generating fake profiles, surveillance), fairness considerations (e.g., deployment of technologies that could make decisions that unfairly impact specific groups), privacy considerations, and security considerations.
        \item The conference expects that many papers will be foundational research and not tied to particular applications, let alone deployments. However, if there is a direct path to any negative applications, the authors should point it out. For example, it is legitimate to point out that an improvement in the quality of generative models could be used to generate Deepfakes for disinformation. On the other hand, it is not needed to point out that a generic algorithm for optimizing neural networks could enable people to train models that generate Deepfakes faster.
        \item The authors should consider possible harms that could arise when the technology is being used as intended and functioning correctly, harms that could arise when the technology is being used as intended but gives incorrect results, and harms following from (intentional or unintentional) misuse of the technology.
        \item If there are negative societal impacts, the authors could also discuss possible mitigation strategies (e.g., gated release of models, providing defenses in addition to attacks, mechanisms for monitoring misuse, mechanisms to monitor how a system learns from feedback over time, improving the efficiency and accessibility of ML).
    \end{itemize}
    
\item {\bf Safeguards}
    \item[] Question: Does the paper describe safeguards that have been put in place for responsible release of data or models that have a high risk for misuse (e.g., pre-trained language models, image generators, or scraped datasets)?
    \item[] Answer: \answerYes{} % Replace by \answerYes{}, \answerNo{}, or \answerNA{}.
    \item[] Justification: We describe the safeguards in Sec.~\ref{sec:paradox}.
    \item[] Guidelines:
    \begin{itemize}
        \item The answer \answerNA{} means that the paper poses no such risks.
        \item Released models that have a high risk for misuse or dual-use should be released with necessary safeguards to allow for controlled use of the model, for example by requiring that users adhere to usage guidelines or restrictions to access the model or implementing safety filters. 
        \item Datasets that have been scraped from the Internet could pose safety risks. The authors should describe how they avoided releasing unsafe images.
        \item We recognize that providing effective safeguards is challenging, and many papers do not require this, but we encourage authors to take this into account and make a best faith effort.
    \end{itemize}

\item {\bf Licenses for existing assets}
    \item[] Question: Are the creators or original owners of assets (e.g., code, data, models), used in the paper, properly credited and are the license and terms of use explicitly mentioned and properly respected?
    \item[] Answer: \answerYes{} % Replace by \answerYes{}, \answerNo{}, or \answerNA{}.
    \item[] Justification: We properly credited and cited licenses of existing assets.
    \item[] Guidelines:
    \begin{itemize}
        \item The answer \answerNA{} means that the paper does not use existing assets.
        \item The authors should cite the original paper that produced the code package or dataset.
        \item The authors should state which version of the asset is used and, if possible, include a URL.
        \item The name of the license (e.g., CC-BY 4.0) should be included for each asset.
        \item For scraped data from a particular source (e.g., website), the copyright and terms of service of that source should be provided.
        \item If assets are released, the license, copyright information, and terms of use in the package should be provided. For popular datasets, \url{paperswithcode.com/datasets} has curated licenses for some datasets. Their licensing guide can help determine the license of a dataset.
        \item For existing datasets that are re-packaged, both the original license and the license of the derived asset (if it has changed) should be provided.
        \item If this information is not available online, the authors are encouraged to reach out to the asset's creators.
    \end{itemize}

\item {\bf New assets}
    \item[] Question: Are new assets introduced in the paper well documented and is the documentation provided alongside the assets?
    \item[] Answer: \answerYes{} % Replace by \answerYes{}, \answerNo{}, or \answerNA{}.
    \item[] Justification: We provided proper documents for our released data.
    \item[] Guidelines:
    \begin{itemize}
        \item The answer \answerNA{} means that the paper does not release new assets.
        \item Researchers should communicate the details of the dataset\slash code\slash model as part of their submissions via structured templates. This includes details about training, license, limitations, etc. 
        \item The paper should discuss whether and how consent was obtained from people whose asset is used.
        \item At submission time, remember to anonymize your assets (if applicable). You can either create an anonymized URL or include an anonymized zip file.
    \end{itemize}

\item {\bf Crowdsourcing and research with human subjects}
    \item[] Question: For crowdsourcing experiments and research with human subjects, does the paper include the full text of instructions given to participants and screenshots, if applicable, as well as details about compensation (if any)? 
    \item[] Answer: \answerNA{} % Replace by \answerYes{}, \answerNo{}, or \answerNA{}.
    \item[] Justification: This paper does not include crowdsourcing experiments and research with human subjects.
    \item[] Guidelines:
    \begin{itemize}
        \item The answer \answerNA{} means that the paper does not involve crowdsourcing nor research with human subjects.
        \item Including this information in the supplemental material is fine, but if the main contribution of the paper involves human subjects, then as much detail as possible should be included in the main paper. 
        \item According to the NeurIPS Code of Ethics, workers involved in data collection, curation, or other labor should be paid at least the minimum wage in the country of the data collector. 
    \end{itemize}

\item {\bf Institutional review board (IRB) approvals or equivalent for research with human subjects}
    \item[] Question: Does the paper describe potential risks incurred by study participants, whether such risks were disclosed to the subjects, and whether Institutional Review Board (IRB) approvals (or an equivalent approval/review based on the requirements of your country or institution) were obtained?
    \item[] Answer: \answerNA{} % Replace by \answerYes{}, \answerNo{}, or \answerNA{}.
    \item[] Justification: The paper does not involve crowdsourcing nor research with human subjects.
    \item[] Guidelines:
    \begin{itemize}
        \item The answer \answerNA{} means that the paper does not involve crowdsourcing nor research with human subjects.
        \item Depending on the country in which research is conducted, IRB approval (or equivalent) may be required for any human subjects research. If you obtained IRB approval, you should clearly state this in the paper. 
        \item We recognize that the procedures for this may vary significantly between institutions and locations, and we expect authors to adhere to the NeurIPS Code of Ethics and the guidelines for their institution. 
        \item For initial submissions, do not include any information that would break anonymity (if applicable), such as the institution conducting the review.
    \end{itemize}

\item {\bf Declaration of LLM usage}
    \item[] Question: Does the paper describe the usage of LLMs if it is an important, original, or non-standard component of the core methods in this research? Note that if the LLM is used only for writing, editing, or formatting purposes and does \emph{not} impact the core methodology, scientific rigor, or originality of the research, declaration is not required.
    %this research? 
    \item[] Answer: \answerYes{} % Replace by \answerYes{}, \answerNo{}, or \answerNA{}.
    \item[] Justification: This study is based on large language models, and the use of such
models is discussed in detail in Sec.~\ref{sec:paradox}.

    \item[] Guidelines:
    \begin{itemize}
        \item The answer \answerNA{} means that the core method development in this research does not involve LLMs as any important, original, or non-standard components.
        \item Please refer to our LLM policy in the NeurIPS handbook for what should or should not be described.
    \end{itemize}

\end{enumerate}